\documentclass[journal]{IEEEtran}
\usepackage{amsmath,amsfonts}
\usepackage{algorithmic}
\usepackage{algorithm}
\usepackage{array}
\usepackage[caption=false,font=footnotesize,labelfont=rm,textfont=rm]{subfig}
\usepackage{textcomp}
\usepackage{stfloats}
\usepackage{url}
\usepackage{verbatim}
\usepackage{graphicx}
\usepackage{cite}
\usepackage{booktabs}
\usepackage{makecell}
\usepackage{multirow}
\usepackage{tabularx}
\usepackage{makecell}
\usepackage{amsmath}
\usepackage{amsfonts}
\usepackage{booktabs}
\usepackage{amssymb}
\usepackage{color}
\usepackage{xcolor}
\definecolor{darkgreen}{rgb}{0.0, 0.8, 0.0}
\usepackage{stfloats}
\usepackage{threeparttable}
\begin{document}
\title{MM-STFlowNet: A Transportation Hub-Oriented Multi-Mode Passenger Flow Prediction Method via Spatial-Temporal Dynamic Graph Modeling}
\author{Ronghui Zhang, Wenbin Xing, Mengran Li, Zihan Wang, Junzhou Chen, Xiaolei Ma,~\IEEEmembership{Senior Member, IEEE,} \newline Zhiyuan Liu,~\IEEEmembership{Senior Member, IEEE}, and Zhengbing He,~\IEEEmembership{Senior Member, IEEE} 
\thanks{Our manuscript was first submitted to IEEE Transactions on Intelligent Transportation Systems on August 11, 2024.}
\thanks{This project is jointly supported by National Natural Science Foundation of China (Nos. 52172350, 51775565), Guangdong Basic and Applied Research Foundation (Nos. 2021B1515120032, 2022B1515120072), Guangzhou Science and Technology Plan Project (Nos. 2024B01W0079, 202206030005), Nansha Key RD Program (No. 2022ZD014), Science and Technology Planning Project of Guangdong Province (No. 2023B1212060029). \textit{(Corresponding author: Junzhou Chen.)}}
\thanks{Ronghui Zhang, Wenbin Xing, Mengran Li, Zihan Wang, and Junzhou Chen are with Guangdong Key Laboratory of Intelligent Transportation System, School of intelligent systems engineering, Sun Yat-sen University, Guangzhou 510275, China (e-mail: zhangrh25@mail.sysu.edu.cn; cliftonspramez886\allowbreak@gmail.com; limr39@mail2\allowbreak.sysu.edu.cn; wangzh579\allowbreak@mail2.sysu.edu.cn; chenjunzhou@mail.sysu.edu.cn).}
\thanks{Xiaolei Ma is with the Key Laboratory of Intelligent Transportation Technology and System, School of Transportation Science and Engineering, Beihang University, Beijing 100191, China (e-mail: xiaolei@buaa.edu.cn).}
\thanks{Zhiyuan Liu is with the Jiangsu Key Laboratory of Urban ITS, Jiangsu Province Collaborative Innovation Center of Modern Urban Traffic Technologies, School of Transportation, Southeast University, Nanjing 210096, China (e-mail: zhiyuanl@seu.edu.cn).}
\thanks{Zhengbing He is with the Senseable City Laboratory, Massachusetts Institute of Technology, Cambridge, MA 02139, USA (e-mail: he.zb\allowbreak@hotmail.com).}}

\maketitle

\begin{abstract}
Accurate and refined passenger flow prediction is essential for optimizing the collaborative management of multiple collection and distribution modes in large-scale transportation hubs. Traditional methods often focus only on the overall passenger volume, neglecting the interdependence between different modes within the hub. To address this limitation, we propose MM-STFlowNet, a comprehensive multi-mode prediction framework grounded in dynamic spatial-temporal graph modeling. Initially, an integrated temporal feature processing strategy is implemented using signal decomposition and convolution techniques to address data spikes and high volatility. Subsequently, we introduce the Spatial-Temporal Dynamic Graph Convolutional Recurrent Network (STDGCRN) to capture detailed spatial-temporal dependencies across multiple traffic modes, enhanced by an adaptive channel attention mechanism. Finally, the self-attention mechanism is applied to incorporate various external factors, further enhancing prediction accuracy. Experiments on a real-world dataset from Guangzhounan Railway Station in China demonstrate that MM-STFlowNet achieves state-of-the-art performance, particularly during peak periods, providing valuable insight for transportation hub management.
\end{abstract}


\section{Introduction}
\begin{figure*}[!t]
\centering
\includegraphics[width=6.0in]{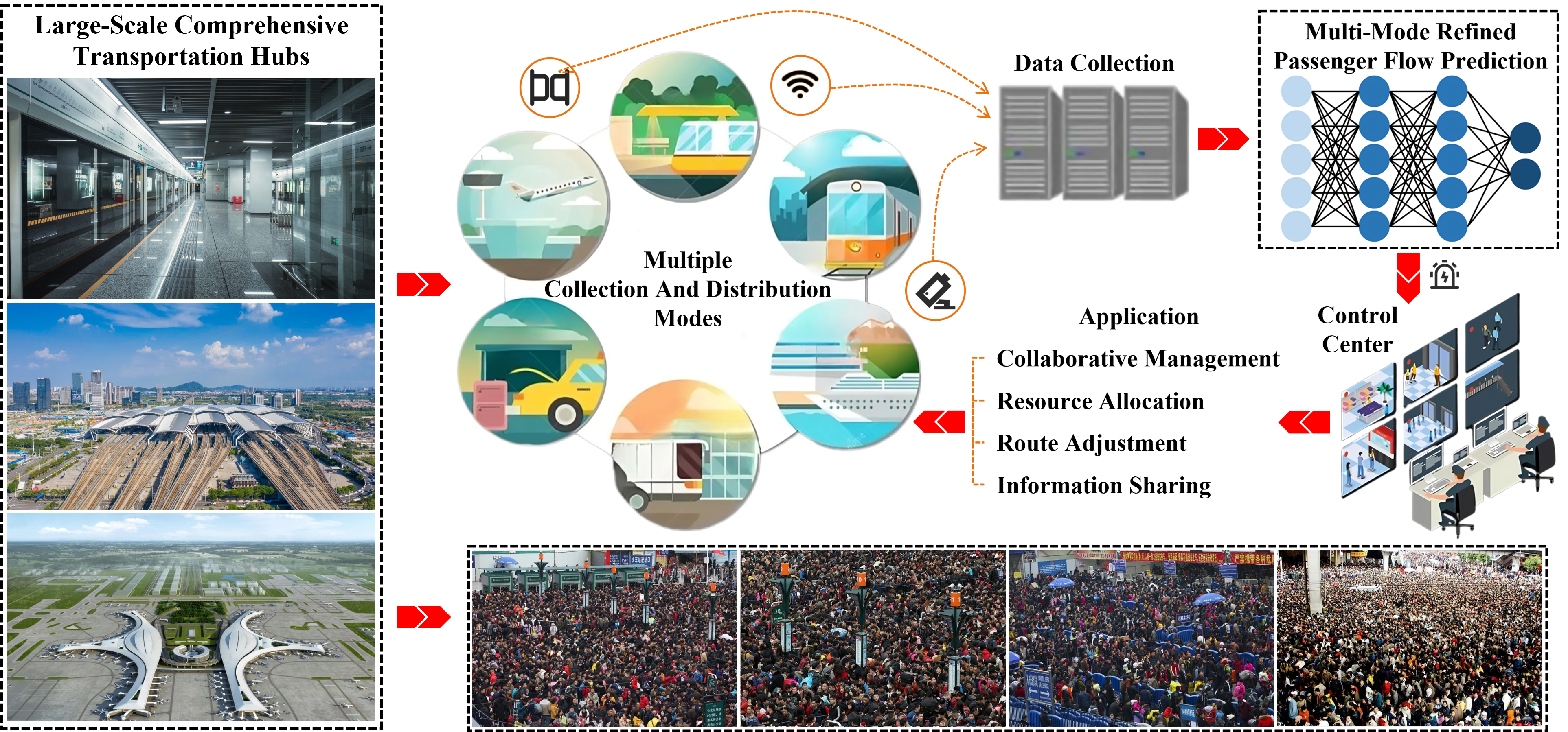}
\caption{Multi-mode refined passenger flow prediction for congestion avoidance. Source: adapted from\cite{pic1,pic2,pic3,pic4}.}
\label{fig:Congestion}
\end{figure*}

\IEEEPARstart{T}{he} rapid urbanization and economic development around the world have led to a significant increase in passenger flow in major transportation hubs in recent years\cite{ghofrani2018recent}. With the flourishing development of intelligent transportation systems (ITS), passengers' travel demands have become increasingly diverse\cite{li2016forecasting}. Multiple collection and distribution modes of hubs present unique challenges and opportunities for transportation planning and management because of the complexity and volume of passenger flow\cite{debrezion2009modelling,wen2012latent}. Accurate passenger flow predictions can provide essential foundational reference for optimizing operational strategies and ensuring the efficient allocation of resources, which is highly valued by operators\cite{zhu2018big}, as shown in Fig. \ref{fig:Congestion}. This has become a critical research area of ITS, with a primary focus on predictions based on historical data\cite{zhang2017real,wang2021metro}. However, achieving precise and practical results remains a challenging task due to various factors\cite{gong2020potential}.

Passengers make individual travel choices based on convenience, cost, and time at different periods. The dynamics and complexity result in completely different and highly volatile temporal patterns for various traffic modes. And traditional prediction methods often focus on the overall trend within the hub, neglecting the interdependence and interaction between different transportation modes. Moreover, major holidays and inclement weather significantly affect passenger flow patterns, leading to localized congestion and increased demand for specific services. Considering these external factors can better achieve the prediction effect. Therefore, there is a pressing need for a comprehensive approach that can capture the intricate dynamics of multi-mode passenger flows.

Traditional methods limit passenger flow prediction to the temporal perspective, typically relying on the Recurrent Neural Network (RNN)\cite{rumelhart1986learning} or Transformer\cite{vaswani2017attention}. Some studies use the Graph Convolutional Network (GCN)\cite{kipf2016semi} to model the spatial attributes of different areas of traffic hubs. However, spatial heterogeneity and intricate interactions between different modes have not been fully explored. And classical spatial distance modeling methods cannot effectively cope with the complex dynamics of the transportation system. It is important to effectively integrate spatial information at different scales to capture traffic mobility within transport hubs. A deeper understanding into the dynamic spatial-temporal correlation between different modes requires to be provided to break the current limitations\cite{li2021spatial,liu2020dynamic,fang2021spatial}. 

\IEEEpubidadjcol
To effectively address these challenges, this study proposes an innovative prediction method, called MM-STFlowNet, to capture the complex interaction between different modes based on the spatial-temporal dynamic traffic graph structure. 

Our contributions can be summarized as follows:
\begin{enumerate}
\item{We present a novel  approach to predict the distribution of passenger flow across various collection and distribution modes in traffic hubs. By emphasizing the necessity of fine-grained analysis, we propose MM-STFlowNet, a refined prediction method that focuses on the dynamic spatial-temporal dependencies of multiple modes while incorporating external factors. This unique perspective offers enhanced practical value for significantly improving management under abnormal conditions.}

\item{To tackle the common issue of inadequate handling of temporal fluctuations and data spikes in current methods, this paper employs an integrated temporal enhancement strategy using signal decomposition and convolution processing techniques. This includes series decomposition to reveal complex temporal patterns, historical enhancement to mitigate data volatility, and peak enhancement to emphasize abnormal data.}

\item{Considering the inherent limitations of static spatial feature extraction in existing studies, we construct the data-driven Spatial-Temporal Dynamic Graph Convolutional Recurrent Network (STDGCRN) based on the multi-mode traffic network. This method comprehensively captures both global and local spatial-temporal patterns, and synthesizes superior feature representations for various prediction tasks by employing an adaptive channel attention mechanism.}

\item{We devise a volatility-enhanced exponential loss function to more accurately capture data variations. Extensive experiments and visual analyses on a real-world dataset demonstrate that our proposed method achieves state-of-the-art prediction performance, particularly during peak periods and holidays. These findings offer valuable insights for enhancing the service and management levels of transportation hubs.}
\end{enumerate}

The remainder is arranged as follows. Section \ref{sec:Literature Review} reviews the related work. Some definitions are provided in Section \ref{sec:Preliminaries}. In Section \ref{sec:Methodology}, we discuss the construction process and motivation of MM-STFlowNet. The details of the experiment and the analysis are shown in Section \ref{sec:Experiments and Analysis}. Finally, the conclusions and future development are given in Section \ref{sec:Conclusion and Future Development}.

\section{Literature Review}
\label{sec:Literature Review}
This section first reviews relevant passenger flow prediction methods for transportation hubs based on deep learning. Then, a detailed analysis of spatial-temporal feature extraction and modeling based on the multi-mode traffic network is provided.

\subsection{Deep Learning Based Passenger Flow Forecasting}
As the most typical traffic center, the railway station accommodates all types of passengers, including long-distance travelers, commuters, and temporary passengers. Train schedules also vary significantly, resulting in substantial fluctuations in passenger flow. Consequently, accurately forecasting passenger flow at railway stations is more challenging.
\begin{figure*}[bp]
    \centering
    \subfloat[Dynamic spatial dependencies]{\includegraphics[width=3.4in]{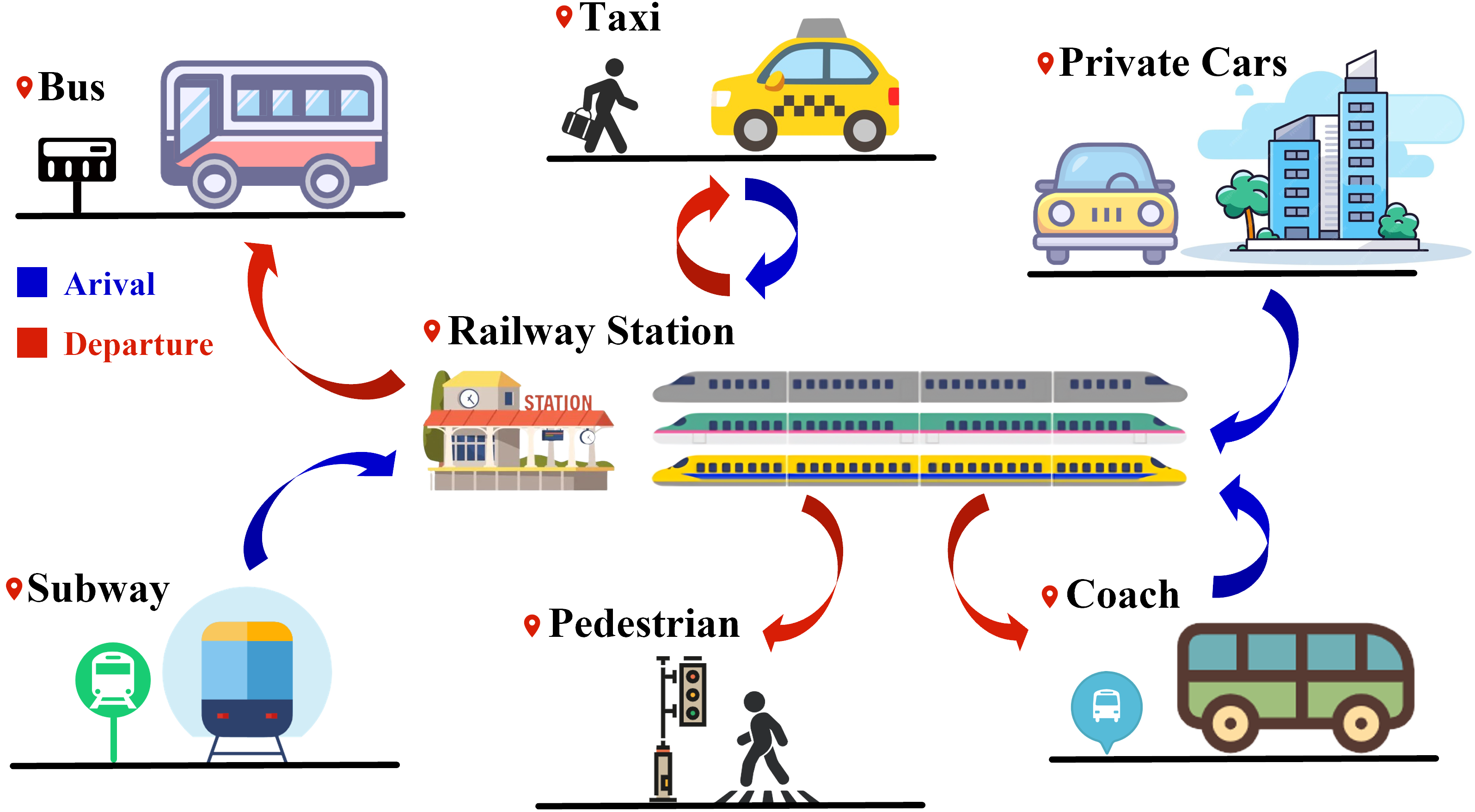}\label{fig:Spatial Feature}}
    \hfil
    \subfloat[Complicated temporal relationship]{\includegraphics[width=3.15in]{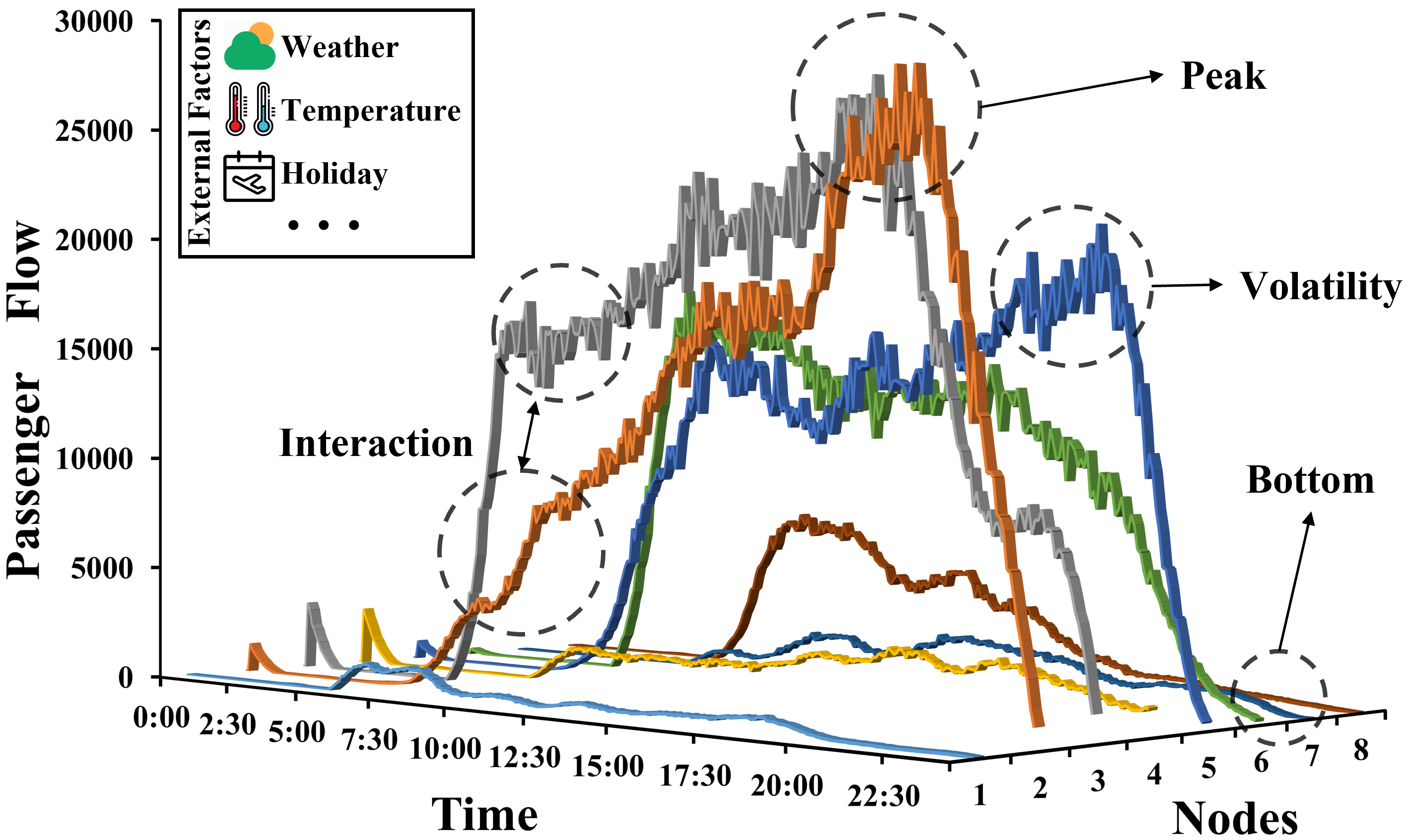}\label{fig:Temporal Feature}}
    \caption{Analysis of spatial-temporal feature extraction and modeling based on the
dynamic multi-mode traffic network. Source: adapted from\cite{pic1}.}
    \label{fig:spatial-temporal}
\end{figure*}

Deep learning methods, with their strong feature extracting and nonlinear fitting capabilities, can capture complex traffic patterns and have become the mainstream approach in current passenger flow forecasting at train stations\cite{gong2020potential}. There are three most commonly used approaches for temporal modeling. Traditional RNN\cite{rumelhart1986learning} and its variants such as Long Short-Term Memory (LSTM)\cite{hochreiter1997long} and Gate Recurrent Unit (GRU)\cite{chung2014empirical} are the most typical prediction methods. Although they offered a mechanism to capture temporal dependencies by recursively processing current and previous information\cite{jiang2022deep}, their effectiveness in long-sequence modeling remained limited. The Convolutional Neural Network (CNN) converted passenger flow data into image-like representations to extract the temporal features \cite{guo2019deep}. Using convolution layers can greatly improve the parallel computation capability and efficiency. But the relatively simple models may not fully capture the non-linear relationships and complex patterns in time series data. The self-attention mechanism allows Transformer\cite{vaswani2017attention} to consider all elements in a sequence and establishes various correlations, thereby attracting significant attention from researchers\cite{xie2022multisize,zhang2023cov}. However, the window size limitation affects the context dependency in long sequence processing, impacting training and inference. Moreover, some studies divided railway stations into different zones for predicting passenger flow distribution\cite{dai2024attention}, extracting spatial features based on the GCN\cite{kipf2016semi}.

This paper also analyzes similar studies for other transportation hubs. The subway mainly serves commuters. With a relatively simple traffic pattern, relevant studies made predictions only through smart card data \cite{li2017forecasting,tang2018forecasting}. Some also considered multi-source data, including mobile phone data\cite{chen2019subway} and special events\cite{fu2022short}. Many studies modeled passenger flow at airports, only considering variations in data within airports\cite{murca2018identification,liu2017novel}.

Most existing studies only consider changes in overall passenger flow within the traffic hubs from a temporal perspective, ignoring the impact of other modes of transportation in the surrounding area. As shown in Fig. \ref{fig:spatial-temporal}\subref{fig:Spatial Feature}, passengers arriving at or departing from the train station use various modes. So neglecting the interactions between them can lead to the inability to identify specific traffic patterns. Multi-mode refined forecasting enables managers to understand the specific needs of various traffic modes, thereby improving service quality and effectively identifying potential safety hazards.

\subsection{Spatial-Temporal Feature Extraction and Modeling}
When the interaction of multiple modes is taken into account, the goal becomes to effectively model their complex spatial-temporal correlations\cite{zheng2020gman}.

\subsubsection{The Spatial Dimension}
Passenger flows exhibit significant interdependence and transitivity between different traffic entities, as shown in Fig. \ref{fig:spatial-temporal}\subref{fig:Spatial Feature}. Changes in traffic states alter the connections between different traffic modes, which is rarely fully explored. Building on the successful experience of urban traffic flow forecasting research\cite{medina2022urban}, which extracts spatial features using the distance between different positions, some common methods can be summarized and referenced. CNN was used to capture spatial correlations between local road segments in the road network \cite{zhang2017deep}. A notable limitation is neglecting the irregularity of real-world traffic structures, leading to the loss of critical topological information. A better approach to avoid this problem is to construct traffic networks as graphs \cite{geng2019spatiotemporal}. Specifically, GCN typically uses adjacency or Laplacian matrices to describe the graph structure\cite{kipf2016semi}. Researchers used prior knowledge\cite{li2017diffusion,yu2017spatio}, such as spatial distance or POI similarity, to construct graph structures directly or as a predefined basis, followed by adaptive adjustments\cite{bai2020adaptive}.

Similarly, the multiple modes around the transportation hubs are irregularly distributed, and their spatial associations change dynamically over time in a complex manner. A traffic graph using static or adaptive adjacency matrices cannot dynamically reflect the true spatial relationships at each moment. Therefore, constructing a data-driven dynamic spatial graph structure is of practical significance.

\subsubsection{The Temporal Dimension}
Traffic conditions fluctuate cyclically and variably. Thus, the multi-mode passenger flow data exhibits significant volatility and continuous interaction\cite{lv2014traffic}, which can be seen in Fig. \ref{fig:spatial-temporal}\subref{fig:Temporal Feature}. Unlike focusing solely on the overall passenger flow within the hub, refined forecasting is more challenging but can achieve more detailed predictions. Deep learning models can learn regular temporal patterns through continuous training\cite{polson2017deep}. However, unusual situations like holidays require special handling. Some studies \cite{zhang2020passenger} incorporated external factors into model training to adapt to abnormal situations. Factors such as weather, road conditions, and holidays can be integrated into the network along with temporal data for joint learning. And the time interval of passenger flow also affects the difficulty of capturing temporal features. Within a certain range, larger intervals introduce more volatility and lose more details, making accurate future predictions more challenging \cite{chen2020short,xu2023fast}. 

Especially at the train station, the strong volatility and data spikes of the multi-mode passenger flow creates problems for prediction. And the management department has extremely high requirements for the accuracy during peaks, so as to take corresponding emergency measures. A comprehensive temporal feature enhancement solution needs to be specifically designed for better serving the multi-mode prediction.

\section{Preliminaries}
\label{sec:Preliminaries}
This section provides an overview of the fundamental concepts involved in multi-mode passenger flow prediction tasks.

\subsection{Multi-mode Traffic Network}
A multi-mode traffic hub can be represented as a graph network $G=(V, E, A)$, where $V$ is a set of $N$ nodes and $E$ is a set of connected edges. Each mode of transport can generate passenger flow data in up to two directions (i.e., arrival and departure). Each flow is abstracted as a node. The construction of graphs has no clear topology to refer to, and is completely data-driven. The adjacency matrix $A \in \mathbb{R}^{N \times N}$ is constructed to represent the degree of connection between the vertices. 
\begin{figure*}[bp]
\centering
\includegraphics[width=6.2in]{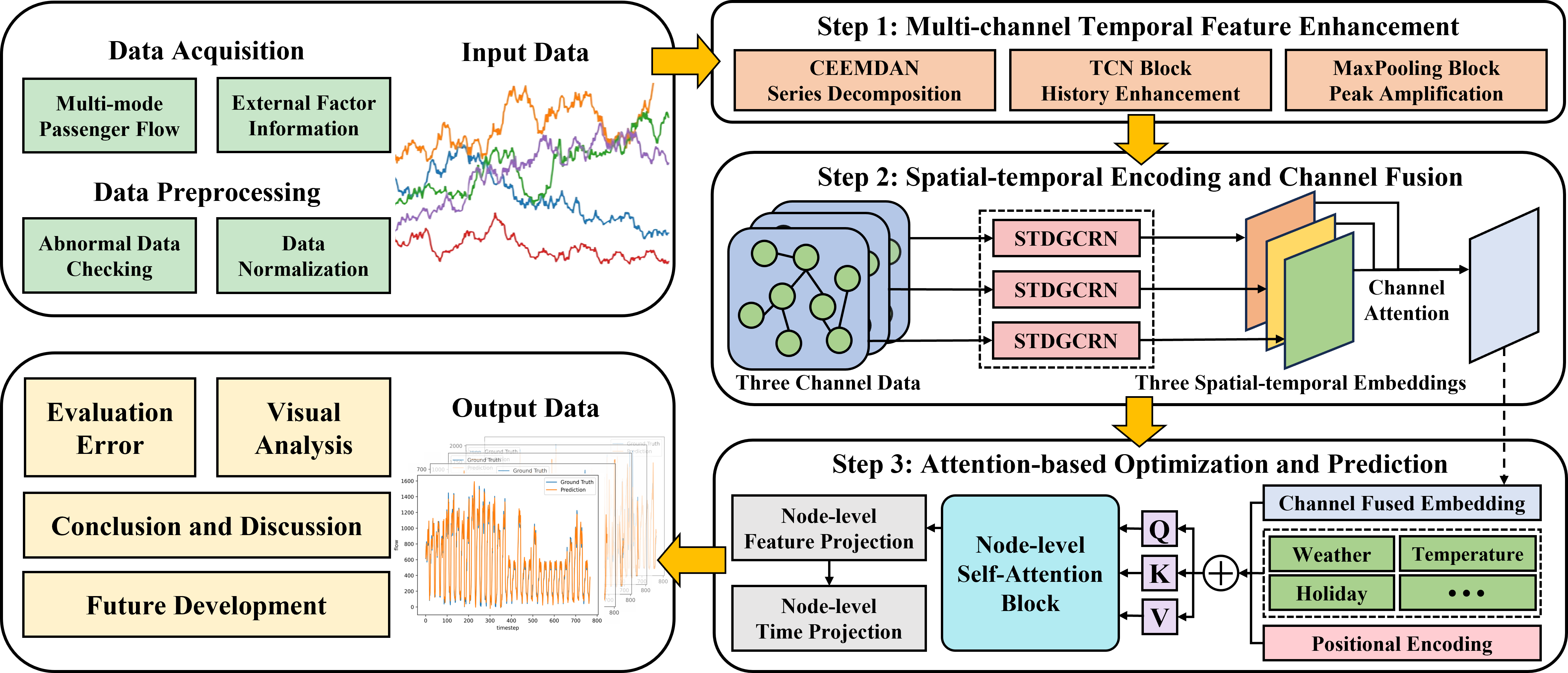}
\caption{The overall proposed framework for refined forecasting of multi-mode passenger flow.}
\label{fig:Framework}
\end{figure*}

\subsection{Multi-mode Passenger Flow Prediction}
The passenger flow data for each node at time step $t$ can be expressed as $x_t\in \mathbb{R}^{N\times C}$, where $C$ represents the number of features. Passenger flow characteristics include volume, speed, density, and so on. Given a fixed historical length of multi-mode passenger flow data $X_P=[x_{t-P+1},\ldots,x_{t-1},x_t]\in \mathbb{R}^{P\times N\times C}$, a complicated function is learned to predict future states $Y_Q=[y_{t+1},\ldots,y_{t+Q}]\in \mathbb{R}^{Q\times N\times C}$, where $P$ represents the size of the history window and $Q$ represents the size of the prediction window. Meanwhile, each 1D sequence can be represented as $s_o\left(t\right)$, where the data flow index $o \leq N$.

\section{Methodology}
\label{sec:Methodology}
This section elaborates on the construction and motivation of the proposed framework, namely MM-STFlowNet.

\subsection{Overview}
As shown in Fig. \ref{fig:Framework}, after the basic processing of the raw data, the first step attempts to enhance the temporal feature on multiple channels. The second step encodes spatial-temporal dependencies synchronously for each channel by creating the novel data-driven spatial-temporal dynamic graph convolutional recurrent network (STDGCRN). Then, the adaptive channel attention mechanism is employed for fusing multi-channel embeddings. The third step applies the self-attention mechanism and incorporates external factors. And the performance is evaluated through error analysis and visual analysis of the predictions against the ground truth. 

\subsection{Multi-channel Temporal Feature Enhancement}
The original $X_p$ data is processed into $D_p$, $B_p$, and $M_p$ by the comprehensive temporal enhancement solution (i.e., series decomposition, history enhancement, and peak amplification). The aim is to deal with strong volatility, data sparsity, and data spikes.

\subsubsection{Series Decomposition}
Empirical Mode Decomposition (EMD) \cite{huang1998empirical} can data-adaptively decompose a signal sequence into a series of Intrinsic Mode Functions (IMFs). Each IMF represents oscillatory components within different frequency ranges \cite{tian2020approach}. The Complete Ensemble Empirical Mode Decomposition with Adaptive Noise (CEEMDAN) \cite{torres2011complete} introduces adaptive noise to address the sensitivity of EMD to noise when processing non-stationary signals, resulting in improved mode decomposition. For each time series, CEEMDAN is applied to capture different temporal patterns and alleviate data volatility.

In a given 1D sequence $s_o\left(t\right)$, Gaussian white noise $z_o^{i}\left(t\right)$ is added $I$ times, followed by the EMD decomposition. The first component $IMF_o^{1}\left(t\right)$ is averaged, shown in Eq. \eqref{Eq:CEEMDAN1}:
\begin{equation}
\label{Eq:CEEMDAN1}
IMF_o^{1}\left(t\right)=\frac{1}{I}\sum_{i=1}^{I}{\operatorname{EMD}\left[s_o\left(t\right)+\varepsilon_o^{i} z_o^{i}\left(t\right)\right]}
\end{equation}
\begin{equation}
\label{Eq:CEEMDAN2}
r_o^{1}\left(t\right)=s_o\left(t\right)-IMF_o^{1}\left(t\right)
\end{equation}where $\varepsilon_o^{i}$ represents the signal-to-noise ratio. And the first residual $r_o^{1}\left(t\right)$ is obtained through Eq. \eqref{Eq:CEEMDAN2}.

Repeat the above steps with the residual signal obtained until the last one $r_o^{u}\left(t\right)$ is a monotonic function or a constant, where $u$ is the final number of decomposition. For multi-mode passenger flow, the maximum number of decomposed IMFs is $m$. Zero-padding is applied to ensure consistency and facilitate subsequent network model processing. The final result can be represented as Eq. \eqref{Eq:CEEMDAN3}.
\begin{equation}
\label{Eq:CEEMDAN3}
s_o\left(t\right)=\sum_{i=1}^{u}{IMF_o^{i}\left(t\right)}+r_o^{u}\left(t\right)+\sum_{i=u+1}^{m}0\left(t\right)
\end{equation}

The $N$ passenger flows and their respective components decomposed by CEEMDAN are recombined into new data $D_p=\left[d_{t-P+1},\ldots,d_{t-1},d_t\right]\in \mathbb{R}^{P\times N\times C_d}$, where $C_d=m+2$, meaning $m$ IMFs, 1 residual and 1 original sequence. 

\subsubsection{History Enhancement}
\begin{figure}[H]
    \vspace{-0.5cm}
    \centering
    \subfloat[Dilated Causal Convolution]{\includegraphics[width=1.81in]{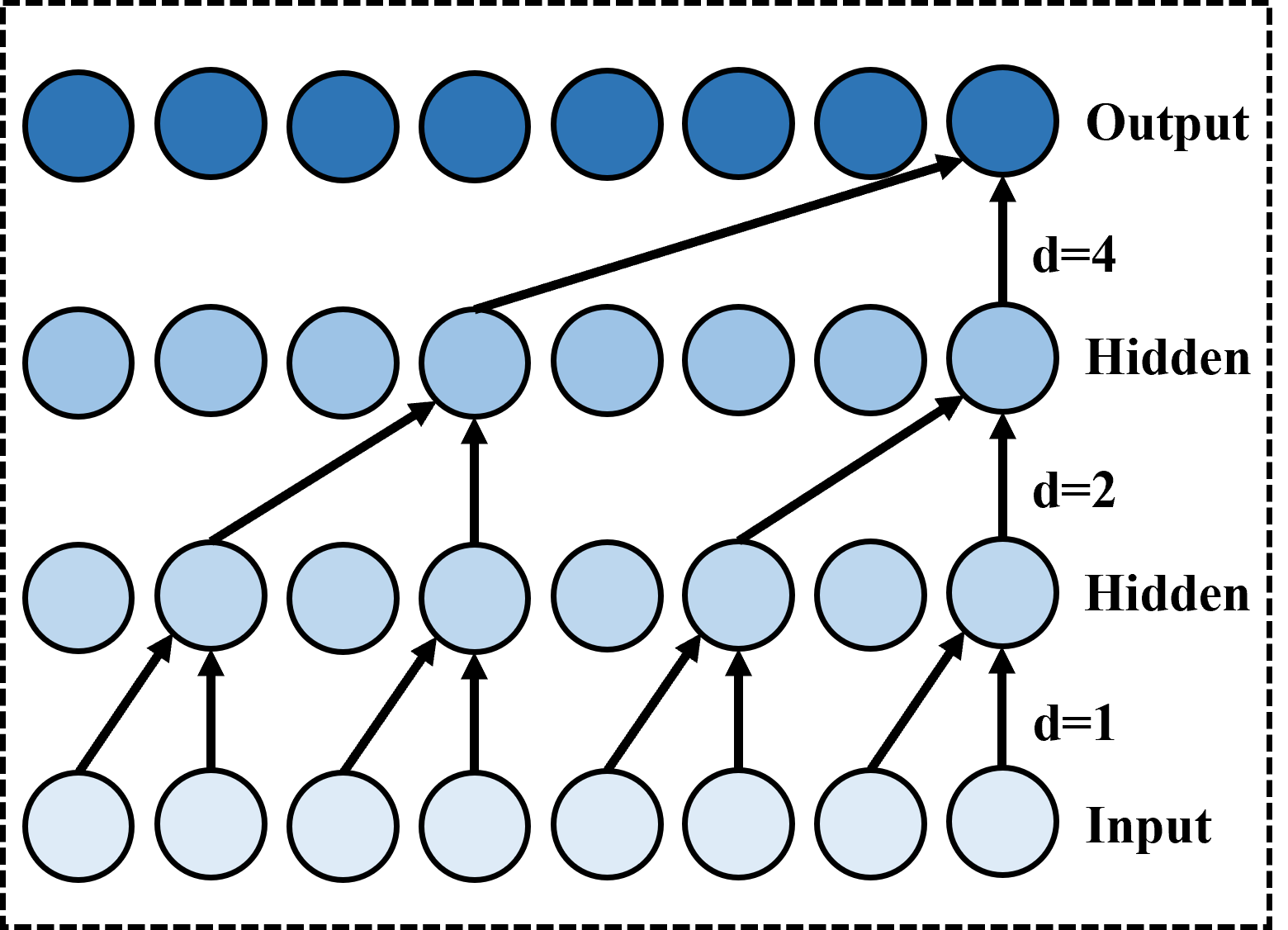}\label{fig:Dilated Causal Conv}}
    \hfil
    \subfloat[TCN Block]{\includegraphics[width=1.4in]{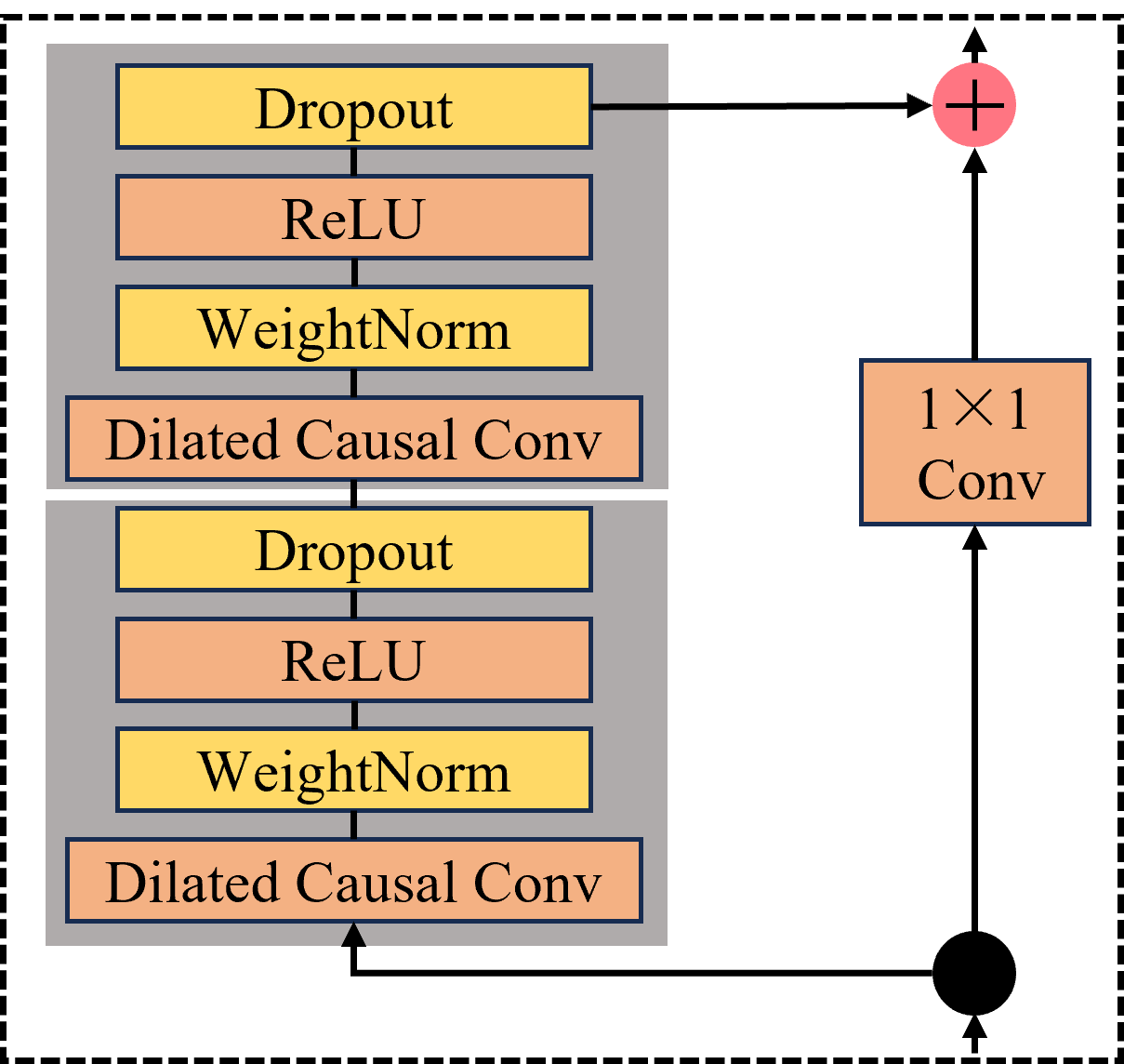}\label{fig:TCN Block}}
    \caption{TCN. (a) displays a dilated causal convolution with dilation factors $d=1, 2, 4$ and filter size $k_T=2$.}
    \label{fig:TCN}
\end{figure}

The Temporal Convolutional Network (TCN)\cite{bai2018empirical} combines the features of dilated convolution and causal convolution. As shown in Fig. \ref{fig:TCN}\subref{fig:Dilated Causal Conv}, dilated convolution introduces a dilation factor into the kernel, allowing it to extract features over a larger receptive field. Causal convolution, on the other hand, uses a strictly time-constrained way that ensures each kernel can only depend on its preceding input data. Additionally, by combining dilated causal convolution with residual connections, the TCN block is obtained in Fig. \ref{fig:TCN}\subref{fig:TCN Block}.

For a given 1D sequence $s_o\left(t\right)$, the dilated causal convolution can be represented by Eq. \eqref{Eq:TCN1}:
\begin{equation}
\label{Eq:TCN1}
\operatorname{DConv}(s_o\left(t\right))=\sum_{i=0}^{k_T-1} s_o(t-d \cdot i) \cdot k_T(i)
\end{equation}where $d$ is the dilated factor and $k_T$ is the kernel size. 

Each dilated causal convolution block can be represented using Eq. \eqref{Eq:TCN2}. The combination of two dilated causal convolution blocks and residual connections forms a TCN block.
\begin{align}
\label{Eq:TCN2}
\operatorname{DCBlock}(s_o(t)) =& \operatorname{Dropout}\Big(\operatorname{ReLU}\big(\operatorname{WeightNorm} \nonumber \\
&\big(\operatorname{DConv}(s_o(t))\big)\big)\Big)
\end{align}

Combining TCN blocks with different receptive fields, the ability to transfer long-sequence information is further improved. The $N$ 1D passenger flows and their respective components processed by TCN blocks are recombined into new data $B_p=\left[b_{t-P+1},\ldots,b_{t-1},b_t\right]\in \mathbb{R}^{P\times N\times C_b}$, where $C_b$ represents the number of enhanced channels.

\subsubsection{Peak Amplification}
Combining different MaxPooling 1D kernels can capture peak information on different time scales. Smaller kernel sizes capture finer temporal patterns, whereas larger kernel sizes focus more on the overall temporal structure. The whole process can be shown in Eq. \eqref{Eq:Max}:
\begin{equation}
\label{Eq:Max}
\operatorname{MConv}(s_o\left(t\right))=\sum_{i=0}^{k_M-1} s_o(t) \cdot k_M(i)
\end{equation}where $k_M$ is the kernel size of MaxPooling 1D convolution. The $N$ 1D passenger flows and their respective components amplified are recombined into new data $M_p=\left[m_{t-P+1},\ldots,m_{t-1},m_t\right]\in \mathbb{R}^{P\times N\times C_m}$, where $C_m$ represents the number of amplified channels.

\subsection{Spatial-temporal Encoding and Channel Fusion}
To better integrate spatial-temporal features synchronously, this paper constructs STDGCRN to encode data from three channels separately based on the novel data-driven dynamic graph convolution. This operation can comprehensively handle global and local spatial-temporal associations. Subsequently, an adaptive channel attention mechanism is employed to better fuse the embeddings.

\subsubsection{STDGCRN}
The dynamic graph convolution can be dynamically learned from the node features and structural information of the graph. This paper attempts to extract the spatial-temporal correlation of the nodes from each channel at every time step and design a data-driven method to form a dynamic adjacency matrix.
\begin{figure}[H]
\centering
\includegraphics[width=3.0in]{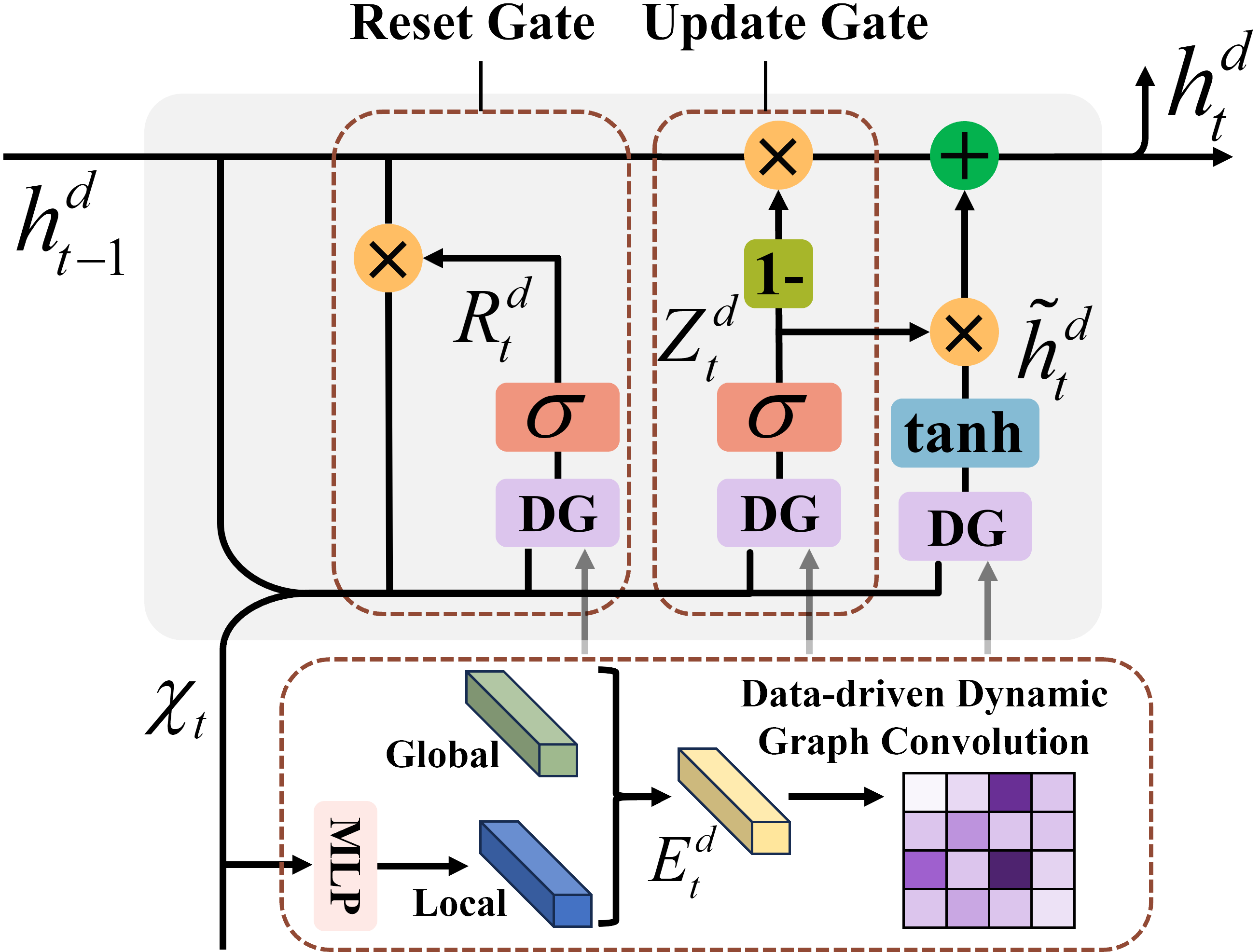}
\caption{STDGCRU. Each memory unit has a data-driven dynamic graph convolution kernel. STDGCRN includes multi-layer STDGCRUs.}
\label{fig:STDGCRU}
\end{figure}

The input at the current time step $t$ is passed through an $\operatorname{MLP}$ layer to extract local dynamic signals. Each node's learnable embedding $E_t\in \mathbb{R}^{N\times L}$ is constructed to represent respective the global spatial-temporal patterns. Subsequently, an element-wise multiplication is performed to generate a dynamic graph embedding $E_t^d\in \mathbb{R}^{N\times L}$, as shown in Eq. \eqref{Eq:STDGCRN1}:
\begin{equation}
\label{Eq:STDGCRN1}
E_t^d=\tanh(E_t\odot \operatorname{MLP}\left(\chi_t\right))
\end{equation}
\begin{equation}
\label{Eq:STDGCRN2}
A_t^d=E_t^dE_t^{d^T}
\end{equation}where $\chi_t$ can be data from each channel (i.e., $d_t$, $b_t$, $m_t$) and $E_t$ is learned independently in each channel. $L$ represents the feature dimension of each channel. The spatial dependence can be inferred according to inter-node similarities by Eq. \eqref{Eq:STDGCRN2}.

To meet the requirements of Chebyshev polynomials\cite{kipf2016semi}, the generated dynamic convolution kernel is normalized as shown in Eq. \eqref{Eq:STDGCRN3}:
\begin{equation}
\label{Eq:STDGCRN3}
g_t^d=I_N+{D_t^d}^{-\frac{1}{2}}\Big(\operatorname{ReLU}\left(A_t^d\right)\Big){D_t^d}^{-\frac{1}{2}}	
\end{equation}
where $g_t^d \in \mathbb{R}^{N\times N}$ and $D_t^d \in \mathbb{R}^{N\times N}$ are the dynamic matrices at time $t$. $I_N$ represents the identity matrix. $D_t^d$ is the degree matrix of $A_t^d$, where the degree of a node represents the number of connected edges. The whole construction can be seen in Fig. \ref{fig:STDGCRU}.

The reset gate selectively determines the amount of information to retain, as shown in Eq. \eqref{Eq:STDGCRN4}, whereas the update gate determines how much information is used for updating, as shown in Eq. \eqref{Eq:STDGCRN5}:
\begin{equation}
\label{Eq:STDGCRN4}
R_t^d=\sigma\left(g_t^d\left[h_{t-1}^d,\chi_t\right]\Theta_R+b_R\right)	
\end{equation}
\begin{equation}
\label{Eq:STDGCRN5}
Z_t^d=\sigma\left(g_t^d\left[h_{t-1}^d,\chi_t\right]\Theta_Z+b_Z\right)	
\end{equation}
\begin{equation}
\label{Eq:STDGCRN6}
\tilde{h_t^d}=\tanh{\left(g_t^d\left[R_t^d\odot h_{t-1}^d,x_t\right]\Theta_h+b_h\right)}
\end{equation}where $\mathbb{R}_t^d\in \mathbb{R}^{N\times F}$ and $Z_t^d\in \mathbb{R}^{N\times F}$. $F$ indicates the size of the embedding dimension. The information of the current memory can be calculated through Eq. \eqref{Eq:STDGCRN6}. $\Theta_R$, $\Theta_Z$, $\Theta_h$ and $b_R$, $b_Z$, $b_h$ are the corresponding learnable parameters.

The final content to be retained is obtained by Eq. \eqref{Eq:STDGCRN7}, where $h_t^d\in \mathbb{R}^{N\times F}$, as shown in Fig. \ref{fig:STDGCRU}. The spatial-temporal features of each channel are learned independently at each time step using a dedicated spatial-temporal dynamic graph convolutional recurrent unit (STDGCRU), resulting in a layer of $P$ STDGCRUs corresponding to the fixed historical length $P$. STDGCRN is essentially multiple layers of STDGCRUs.
\begin{equation}
\label{Eq:STDGCRN7}
h_t^d=Z_t^d\odot\tilde{h_t^d}+\left(1-Z_t^d\right)\odot h_{t-1}^d
\end{equation}

The final multi-channel spatial-temporal representations (i.e., $\hat{D_p}, \hat{B_p}, \hat{M_p}\in \mathbb{R}^{P\times N\times F}$) are obtained, indicating that each encoding channel is learned separately and updates its parameters according to the distribution characteristics of its own data.

\subsubsection{Channel Attention Mechanism}
\begin{figure}[H]
\centering
\includegraphics[width=3.0in]{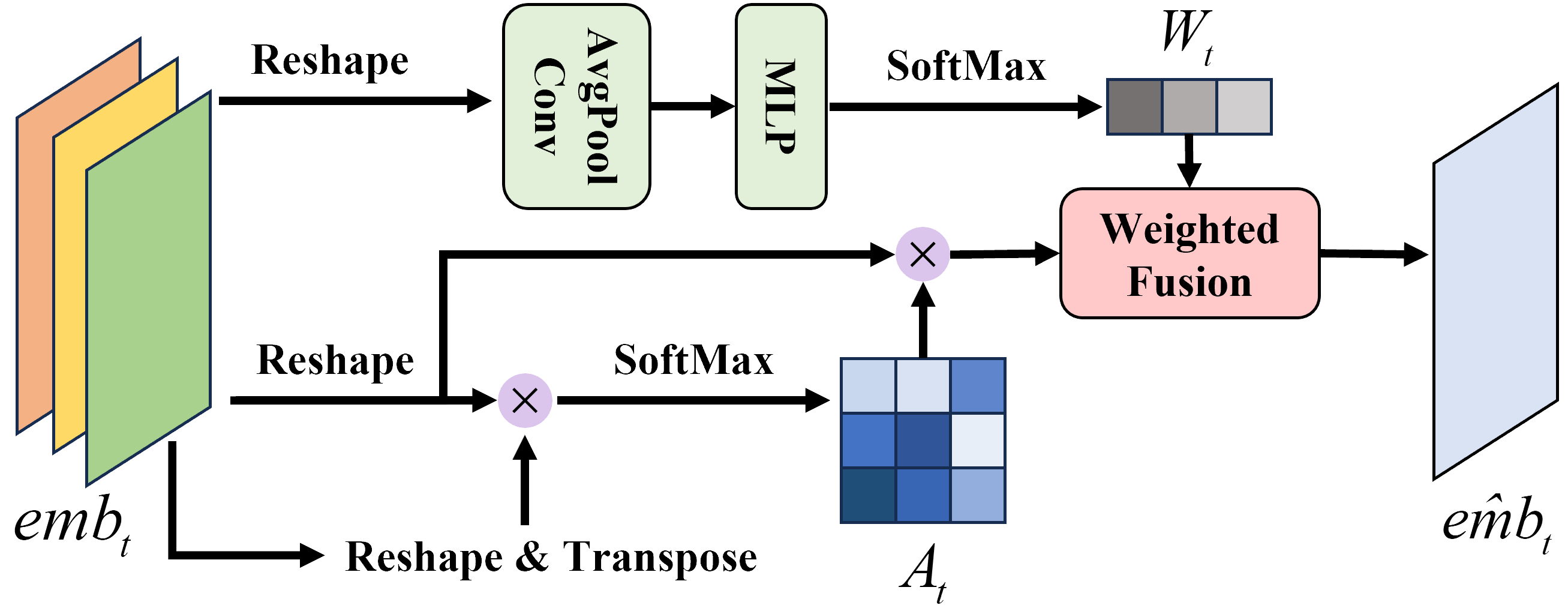}
\caption{The adaptive channel attention mechanism.}
\label{fig:Channel Attention}
\end{figure}

The adaptive channel attention mechanism designed can improve the representation ability of features, and pay more attention to the important information for different downstream prediction tasks. The whole computing process is displayed in Fig. \ref{fig:Channel Attention}. 

At each time step $t$, multi-channel embeddings are stacked into a single one, i.e., $emb_t\in \mathbb{R}^{3\times N\times F}$. The shape of $emb_t$ is then transformed to obtain $\bar{emb_t}\in \mathbb{R}^{3\times(N\times F)}$ to obtain the channel attention map $A_t\in R^{3\times3}$, as represented in Eq. \eqref{Eq:CA3}:
\begin{equation}
\label{Eq:CA2}
\alpha=\sqrt{N\times F}
\end{equation}
\begin{equation}
\label{Eq:CA3}
AM_t=\operatorname{SoftMax}\big((\bar{emb_t}/\alpha)\cdot\bar{{emb_t}}^T\big)
\end{equation}where $\alpha$ is a scale factor, computed as Eq. \eqref{Eq:CA2}.

The AvgPooling layer is used to extract the overall weights $W_t$, as shown in Eq. \eqref{Eq:CA4}. Subsequently, the multi-channel spatial-temporal embeddings are strengthened by using the channel attention map $AM_t$ to obtain $\tilde{emb_t}\in \mathbb{R}^{3\times\left(N\times F\right)}$ through Eq. \eqref{Eq:CA5}.
\begin{equation}
\label{Eq:CA4}
W_t=\operatorname{SoftMax}\Big(\operatorname{MLP}\big(\operatorname{AvgPooling}\left(emb_t\right)\big)\Big)
\end{equation}
\begin{equation}
\label{Eq:CA5}
\tilde{emb_t}=AM_t\cdot\bar{emb_t}
\end{equation}

The shape of $\tilde{emb_t}$ is adjusted to obtain $\breve{emb_t}\in \mathbb{R}^{3\times N\times F}$, and then the final representation $\hat{emb_t}\in \mathbb{R}^{N\times F}$ is merged by the overall weight $W_t$, as shown in Eq. \eqref{Eq:CA6}. The channel attention mechanism increases sensitivity to key features. Eventually, spatial-temporal embeddings $\hat{emb_p}\in \mathbb{R}^{P\times N\times F}$ are obtained.
\begin{equation}
\label{Eq:CA6}
\hat{emb_t}=\sum_{i=1}^{3}\left({W_t}_\mathrm{i}\odot{\breve{emb_t}}_i\right)
\end{equation}

\subsection{Attention-based Optimization and Prediction}
In order to further enhance the key features distributed in different time steps and the prior guidance effect of various environmental conditions, this article applies the self-attention mechanism that integrates multiple external factors. 

A given sequence is mapped into a query vector $Q$, a key vector $K$ of the same dimension $d_k$, and a value vector $V$ of dimension $d_v$. The self-attention mechanism\cite{vaswani2017attention} is shown in Eq. \eqref{Eq:self-attention} below:
\begin{equation}
\label{Eq:self-attention}
\operatorname{Self Attention}(Q, K, V)=\operatorname{Softmax}\left(\frac{Q K^{\top}}{\sqrt{d_k}}\right) V
\end{equation}

All external information is converted into numeric types and encoded into a learnable continuous vector representation. And position encoding is generally composed of a sine function and a cosine function, which can provide the model with position information at different frequencies. The spatial-temporal, positional and external factor embeddings are added together to obtain the final representation ${Emb}_P\in \mathbb{R}^{P\times N\times F}$, which serves as input for the self-attention block. For each representation, a self-attention block is used for temporal attention enhancement using Eqs. \eqref{Eq:SA4} and \eqref{Eq:SA5}:
\begin{equation}
\label{Eq:SA4}
\hat{{Emb}_P}=\operatorname{LayerNorm}({Emb}_P)
\end{equation}
\begin{align}
\label{Eq:SA5}
{\breve{{Emb}_P}}_i =& \operatorname{Self Attention}\big(\operatorname{Dropout}(\hat{{Emb}_P}_i, \nonumber \\
&\hat{{Emb}_P}_i,\hat{{Emb}_P}_i )\big)+\hat{{Emb}_P}_i
\end{align}where $i=1,\ldots,N$, indicating the serial number of the multiple passenger flows, and the final embeddings $\breve{{Emb}_P}\in \mathbb{R}^{P\times N\times F}$.

Finally, the embeddings of each node in the time dimension and feature dimension are projected respectively to obtain the prediction data $\hat{Y_Q}=\left[\hat{y_{t+1}},\ldots,\hat{y_{t+Q}}\right]\in \mathbb{R}^{Q\times N\times C}$. And each 1D predicted sequence can be represented as $\hat{s_o\left(t\right)}$.

\section{Experiments and Analysis}
\label{sec:Experiments and Analysis}
In this section, MM-STFlowNet is evaluated and compared on a multi-mode dataset of passenger flow collected at Guangzhounan Railway Station in China. This paper selects three typical passenger flow patterns for visual analysis. In order to obtain the best prediction effect, the optimal historical window size is also explored. In addition, a detailed ablation experiment is carried out to confirm the effectiveness of each module.

\subsection{Datasets and Preprocessing}
Guangzhounan Railway Station, located in Guangdong Province, is the largest and busiest railway station in China\cite{li2022prediction}. The dataset (GZN) records 11 types of arrival or departure passenger flow data per hour from 0:00 on January 21, 2023 to 23:00 on January 20, 2024, including six modes of transportation: taxi, coach, subway, private car, online ride-hailing, and coach. Each passenger flow is abstractly defined as a node in the traffic graph network. And passenger flow arriving and leaving by train can be obtained directly from the schedule and ticket purchase information in advance. Therefore, there is no need for relevant data by trains in the dataset. The details of GZN are shown in Table \ref{tab:dataset}. 
\begin{table}[H]
\caption{Statistics of Two Datasets\label{tab:dataset}}
\centering
\resizebox{\columnwidth}{!}{%
\begin{tabular}{ccccccc}
\toprule
Dataset & Time Steps & Nodes & Interval & Max   & Min & Mean \\ \midrule
GZN      & 8760       & 11    & 1 h      & 28196 & 0   & 1341 \\ 
Traffic  & 17544      & 862   & 1 h      & 0.724 & 0   & 0.057 \\ \bottomrule
\end{tabular}%
}
\end{table}

\begin{figure}[H]   
\centering
\includegraphics[width=3.2in]{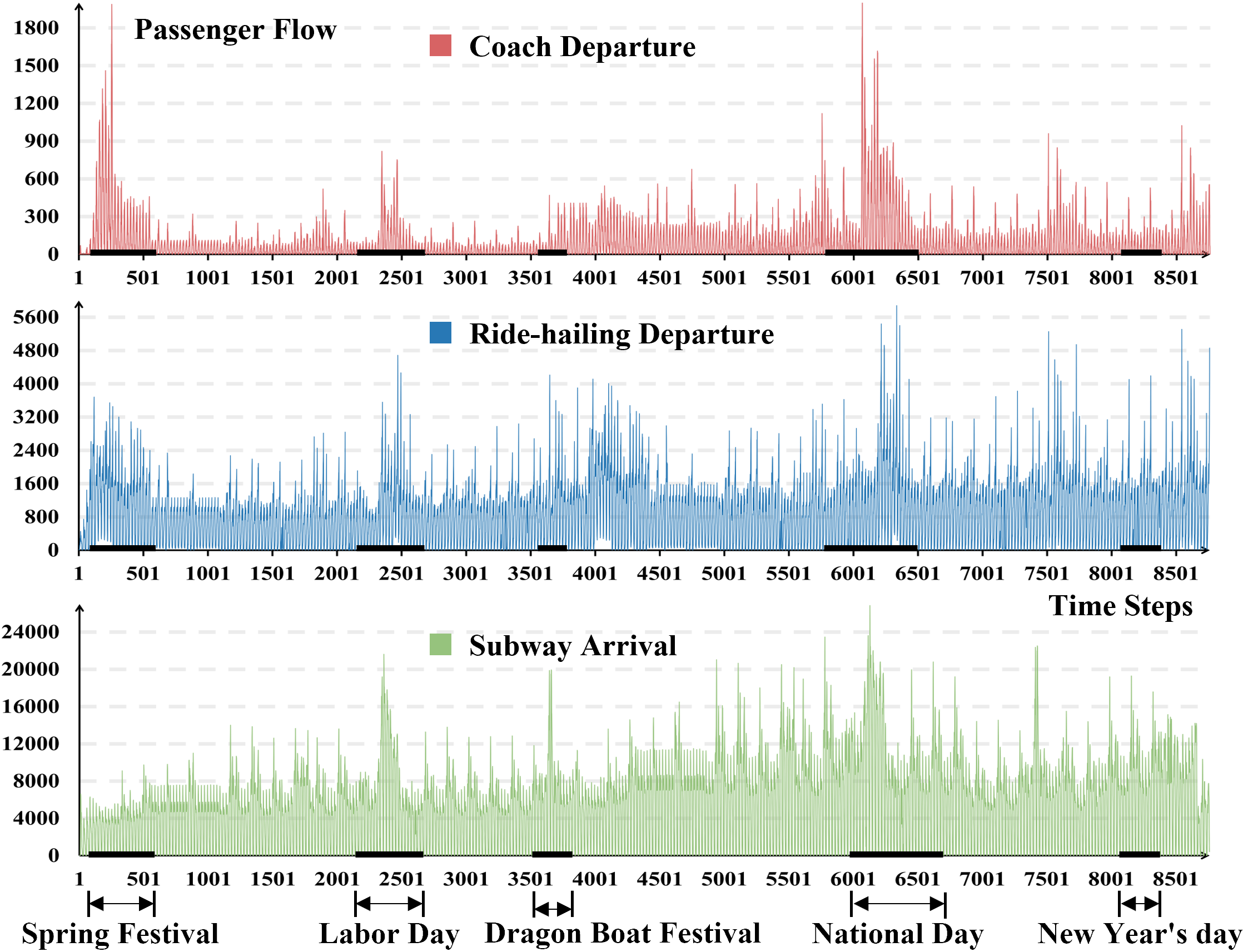}
\caption{Comparison of temporal patterns among three typical passenger flows. The time periods of major holidays are marked in bold on each horizontal axis, and the corresponding time range is displayed in Table \ref{tab:holiday}.}
\label{fig:Data Display}
\end{figure}

Taking the passenger flow of coach departure, ride-hailing departure and subway arrival as examples, as shown in Fig. \ref{fig:Data Display}, there are significant differences in temporal patterns. The passenger flow of coaches has significant sharp drops. Due to the need to carry luggage, people prefer to travel taking a ride-hailing or subway ride. The average number of people arriving at the train station by subway is higher than the peak number arriving by coach. Even during the holidays in Table \ref{tab:holiday}, they all have completely different characteristics. But they all exhibit strong volatility and prominent data spikes. The external information includes seven factors: hour, temperature, wind speed, humidity, visibility, weather, and holidays. Then 80$\%$ of the dataset is taken as the train set and 20$\%$ as the test set.

In order to prove the generalization of MM-STFlowNet, another dataset (Traffic) is selected from the urban road traffic. Traffic is a collection of 48 months (2015-2016) hourly data from the California Department of Transportation. It describes the road occupancy rates measured by different sensors on San Francisco Bay area freeways \cite{lai2018modeling}. The details of Traffic are shown in Table \ref{tab:dataset}.

Data normalization helps accelerate the convergence and improves the performance and stability. Therefore, we use the Min-Max normalization method is adopted on each passenger flow.
\begin{table}[H]
\caption{Statistics of The Five Major Chinese Legal Holidays\label{tab:holiday}}
\centering
\begin{tabular}{ccccc}
\toprule
Holiday            & Date Range \\ \midrule
Spring Festival    & Jan. 21-27, 2023 \\
Labor Day         & Apr. 29-May 3, 2023 \\
Dragon Boat Festival & Jun. 22-24, 2023 \\
National Day       & Sep. 29-Oct. 6, 2023 \\
New Year's Day     & Dec. 30, 2023-Jan. 1, 2024 \\ \bottomrule
\end{tabular}
\end{table}

\subsection{Loss Function}
In regression problems, common loss functions include the mean absolute error (MAE) loss, the mean squared error (MSE) loss, and the quantile loss, which can be represented from Eq. \ref{Eq:MAE} to Eq. \ref{Eq:Quantile}, respectively. $\tau$ is the quantile level, usually with a value between 0 and 1.
\begin{equation}
\label{Eq:MAE}
\operatorname{MAE}(s_o\left(i\right), \hat{s_o\left(i\right)})=|s_o\left(i\right)- \hat{s_o\left(i\right)}|
\end{equation}
\begin{equation}
\label{Eq:MSE}
\operatorname{MSE}(s_o\left(i\right), \hat{s_o\left(i\right)})=(s_o\left(i\right)- \hat{s_o\left(i\right)})^2
\end{equation}
\begin{align}
\label{Eq:Quantile}
\operatorname{Quantile}&(s_o(i), \hat{s_o}(i)) = \nonumber \\
& \begin{cases} 
(1-\tau) \cdot(s_o(i)-\hat{s_o}(i)) & \text{if } s_o(i) > \hat{s_o}(i) \\
\tau \cdot(s_o(i)-\hat{s_o}(i)) & \text{if } s_o(i) \leq \hat{s_o}(i)
\end{cases}
\end{align}

MAE penalizes error linearly and is therefore insensitive to outliers. Due to the square term, MSE is more sensitive to large errors and can more effectively punish outliers. The quantile loss controls the penalty for error in different directions by adjusting $\tau$. 
\begin{figure}[H]   
\centering
\includegraphics[width=2.5in]{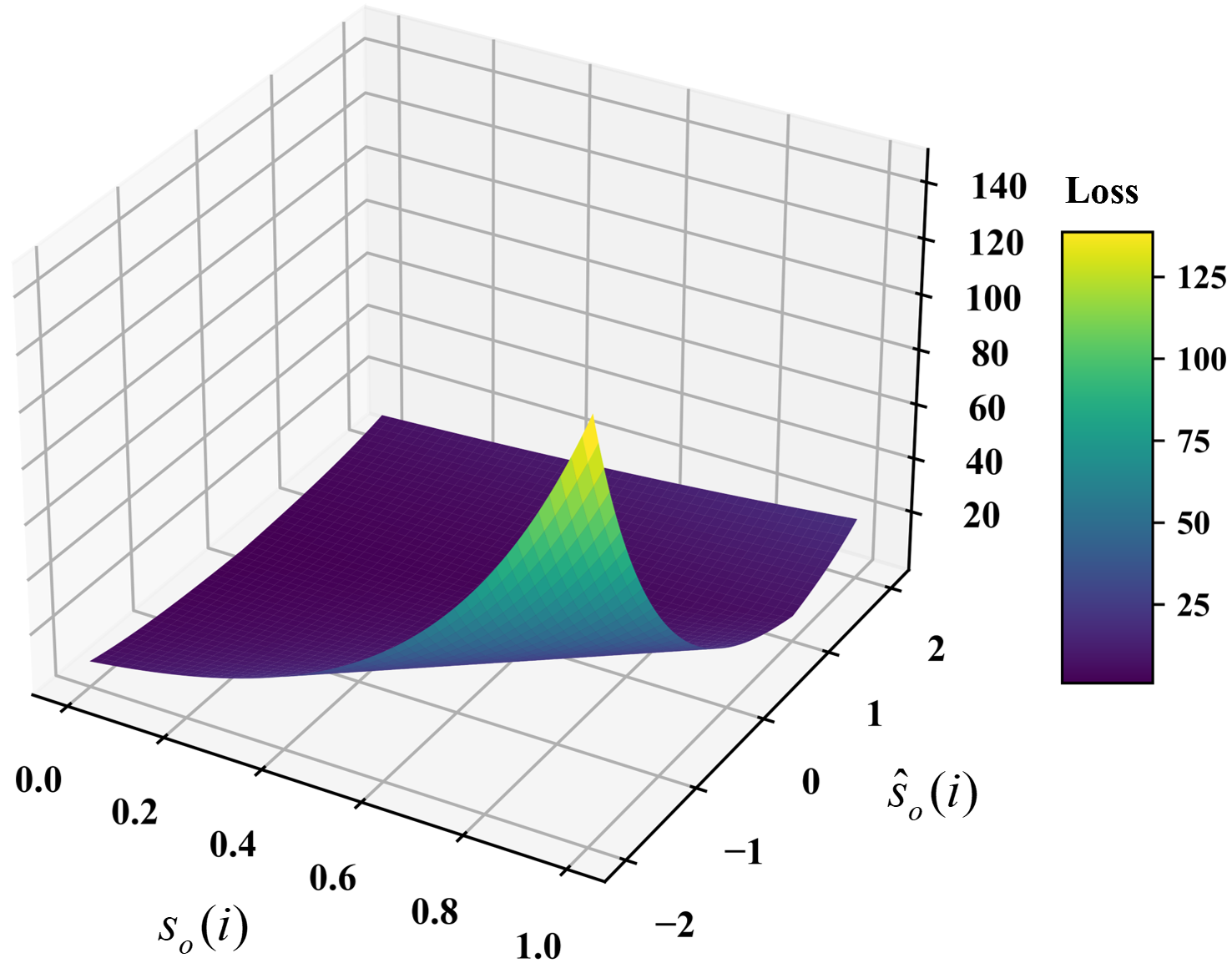}
\caption{EPEL function. $p=2$ and $q=1$.}
\label{fig:EPEL}
\end{figure}

Based on these loss functions, in order to improve the prediction accuracy of peak data, this paper designs a novel loss function called Enhanced Peak Exponential Loss (EPEL) as shown in Eq. \ref{Eq:EPEL}:
\begin{equation}
\label{Eq:EPEL}
\operatorname{EPEL}(s_o(i), \hat{s_o}(i))=\left(e^{s_o(i)}\right)^p \cdot \left(e^{|s_o\left(i\right)- \hat{s_o\left(i\right)}|}\right)^q
\end{equation}where $p$ and $q$ are the adjustment terms that control the sensitivity of EPEL to different errors. The data is scaled to $[0,1]$ after normalization. Therefore, the data is mapped to $[1,\infty]$ with exponential amplification. The error is further amplified by adjusting $p$ and $q$, and a higher penalty is imposed on the error of the peak data. As shown in Fig. \ref{fig:EPEL}, the exponential functions and different exponential terms makes it more sensitive to larger errors, especially for peak. 
\begin{table*}[bp]
\centering
\caption{Performance Comparison with Different Prediction Window Sizes on GZN}
\label{tab:Comparison GZN}
\small
\scriptsize 
\belowrulesep=0pt
\aboverulesep=0pt
\setlength{\tabcolsep}{3.5pt}
\begin{tabularx}{\textwidth}{ccc|ccccccccccc}
\toprule
\multirow{3}{*}{\makecell[c]{Predictive\\Size}} & \multirow{3}{*}{Period} & \multirow{3}{*}{Metrics} & \multicolumn{2}{c|}{RNN-based} & \multicolumn{3}{c|}{Transformer-based} & \multicolumn{3}{c|}{Graph-based} & \multicolumn{1}{c|}{CNN-based} & \multicolumn{1}{c|}{Multi-scale} & \multirow{3}{*}{\textbf{Ours}}\\
\cmidrule(r){4-13}  \cmidrule(lr{0pt}){12-13} 
 &  &  & \multicolumn{1}{c}{LSTM} & \multicolumn{1}{c|}{GRU} & \makecell[c]{Transformer\\NeurIPS'2017} & \makecell[c]{Pyraformer\\ICLR'2022} &  \multicolumn{1}{c|}{\makecell[c]{iTransformer\\ICLR'2024}}  & \makecell[c]{ASTGCN\\AAAI'2019} & \makecell[c]{ST-GDN\\AAAI'2021} & \multicolumn{1}{c|}{\makecell[c]{Dstagnn\\ICML'2022}} & \multicolumn{1}{c|}{\makecell[c]{TimesNet\\ICLR'2023}} & \multicolumn{1}{c|}{\makecell[c]{TimeMixer\\ICLR'2024}} &  \\
\midrule
\multirow{8}{*}{1} & \multirow{2}{*}{Entire Day} & MSE & 0.0072 & 0.0048 & 0.0033 & 0.0025 & 0.0028 & 0.0042 & 0.0026 & \textcolor{darkgreen}{\textbf{0.0023}} & 0.0024 & \textcolor{darkgreen}{\textbf{0.0023}} & \textcolor{blue}{$\textbf{0.0012}\uparrow$}\\ 
             &      & MAE & 0.0710 & 0.0473 & 0.0379 & 0.0303 & 0.0328 & 0.0451 & \textcolor{darkgreen}{\textbf{0.0231}} & 0.0286 & 0.0299 & 0.0285 & \textcolor{blue}{$\textbf{0.0222}\uparrow$} \\ \cmidrule(r){2-14}
&  \multirow{2}{*}{Evening} & MSE & 0.0089 & 0.0059 & 0.0025 & 0.0030 & 0.0026 & 0.0052 & \textcolor{darkgreen}{\textbf{0.0018}} & 0.0040 & \textcolor{blue}{\textbf{0.0014}} & \textcolor{blue}{\textbf{0.0014}} & \textcolor{blue}{\textbf{0.0014}} \\
             &       & MAE & 0.0872 & 0.0581 & 0.0314 & 0.0373 & 0.0335 & 0.0555 & 0.0220 & 0.0445 & \textcolor{darkgreen}{\textbf{0.0199}} & 0.0264 & \textcolor{blue}{$\textbf{0.0187}\uparrow$} \\ \cmidrule(r){2-14}
& \multirow{2}{*}{Weekend} & MSE & 0.0095 & 0.0063 & 0.0027 & 0.0032 & \textcolor{darkgreen}{\textbf{0.0018}} & 0.0056 & 0.0021 & 0.0034 & 0.0019 & 0.0019 & \textcolor{blue}{$\textbf{0.0013}\uparrow$} \\
             &       & MAE & 0.0973 & 0.0649 & 0.0362 & 0.0416 & \textcolor{darkgreen}{\textbf{0.0268}} & 0.0619 & 0.0294 & 0.0424 & 0.0279 & 0.0270 & \textcolor{blue}{$\textbf{0.0227}\uparrow$} \\ \cmidrule(r){2-14}
& \multirow{2}{*}{\makecell[c]{New\\Year's Day}} & MSE & 0.0089 & 0.0060 & 0.0026 & 0.0031 & \textcolor{darkgreen}{\textbf{0.0016}} & 0.0053 & 0.0017 & 0.0044 & 0.0017 & \textcolor{darkgreen}{\textbf{0.0016}} & \textcolor{blue}{$\textbf{0.0010}\uparrow$} \\
              &             & MAE & 0.0938 & 0.0626 & 0.0358 & 0.0401 & 0.0257 & 0.0597 & 0.0265 & 0.0492 & 0.0256 & \textcolor{darkgreen}{\textbf{0.0252}} & \textcolor{blue}{$\textbf{0.0206}\uparrow$} \\
              \midrule
\multirow{8}{*}{4} & \multirow{2}{*}{Entire Day} & MSE & 0.0128 & 0.0085 & 0.0055 & 0.0036 & 0.0032 & 0.0047 & 0.0045 & \textcolor{darkgreen}{\textbf{0.0031}} & 0.0037 & 0.0054 & \textcolor{blue}{$\textbf{0.0022}\uparrow$} \\ 
               &   & MAE & 0.0945 & 0.0630 & 0.0495 & 0.0378 & 0.0346 & 0.0450 & \textcolor{darkgreen}{\textbf{0.0317}} & 0.0338 & 0.0381 & 0.0468 & \textcolor{blue}{$\textbf{0.0297}\uparrow$} \\ \cmidrule(r){2-14}
 & \multirow{2}{*}{Evening} & MSE & 0.0128 & 0.0085 & 0.0055 & 0.0036 & 0.0032 & 0.0047 & 0.0045 & \textcolor{darkgreen}{\textbf{0.0031}} & 0.0037 & 0.0054 & \textcolor{blue}{$\textbf{0.0022}\uparrow$} \\
               &    & MAE & 0.0945 & 0.0630 & 0.0495 & 0.0378 & 0.0346 & 0.0450 & \textcolor{blue}{\textbf{0.0317}} & 0.0338 & 0.0381 & 0.0468 & \textcolor{darkgreen}{\textbf{0.0324}} \\ \cmidrule(r){2-14}
& \multirow{2}{*}{Weekend} & MSE & 0.0172 & 0.0115 & 0.0041 & 0.0037 & 0.0028 & 0.0063 & 0.0031 & 0.0039 & 0.0028 & \textcolor{darkgreen}{\textbf{0.0027}} & \textcolor{blue}{$\textbf{0.0022}\uparrow$} \\
                &   & MAE & 0.1332 & 0.0888 & 0.0435 & 0.0412 & 0.0336 & 0.0634 & 0.0370 & 0.0432 & 0.0331 & \textcolor{darkgreen}{\textbf{0.0321}} & \textcolor{blue}{$\textbf{0.0296}\uparrow$} \\ \cmidrule(r){2-14}
& \multirow{2}{*}{\makecell[c]{New\\Year's Day}} & MSE & 0.0118 & 0.0079 & 0.0045 & 0.0033 & 0.0027 & 0.0043 & 0.0026 & 0.0041 & 0.0025 & \textcolor{darkgreen}{\textbf{0.0023}} & \textcolor{blue}{$\textbf{0.0021}\uparrow$} \\
                     &     & MAE & 0.1001 & 0.0667 & 0.0469 & 0.0400 & 0.0329 & 0.0476 & 0.0344 & 0.0442 & 0.0314 & \textcolor{darkgreen}{\textbf{0.0304}} & \textcolor{blue}{$\textbf{0.0292}\uparrow$} \\
\midrule          
\multirow{8}{*}{8} & \multirow{2}{*}{Entire Day} & MSE & 0.0089 & 0.0076 & 0.0055 & 0.0047 & 0.0040 & 0.0050 & 0.0053 & \textcolor{darkgreen}{\textbf{0.0038}} & 0.0042 & 0.0060 & \textcolor{blue}{$\textbf{0.0025}\uparrow$} \\ 
                &  & MAE & 0.0711 & 0.0619 & 0.0486 & 0.0422 & 0.0378 & 0.0452 & \textcolor{darkgreen}{\textbf{0.0355}} & 0.0375 & 0.0409 & 0.0496 & \textcolor{blue}{$\textbf{0.0306}\uparrow$} \\ \cmidrule(r){2-14}
 & \multirow{2}{*}{Evening} & MSE & 0.0214 & 0.0196 & 0.0059 & 0.0053 & 0.0043 & 0.0056 & 0.0045 & 0.0052 & 0.0049 & \textcolor{darkgreen}{\textbf{0.0038}} & \textcolor{blue}{$\textbf{0.0023}\uparrow$} \\
                &   & MAE & 0.0932 & 0.0822 & 0.0516 & 0.0473 & 0.0401 & 0.0507 & 0.0431 & 0.0486 & 0.0433 & \textcolor{darkgreen}{\textbf{0.0385}} & \textcolor{blue}{$\textbf{0.0308}\uparrow$} \\ \cmidrule(r){2-14}
& \multirow{2}{*}{Weekend} & MSE & 0.0087 & 0.0078 & 0.0045 & 0.0038 & 0.0033 & 0.0070 & 0.0039 & 0.0039 & 0.0035 & \textcolor{darkgreen}{\textbf{0.0032}} & \textcolor{blue}{$\textbf{0.0026}\uparrow$} \\
                &   & MAE & 0.0757 & 0.0605 & 0.0453 & 0.0408 & 0.0358 & 0.0651 & 0.0406 & 0.0439 & 0.0372 & \textcolor{darkgreen}{\textbf{0.0354}} & \textcolor{blue}{$\textbf{0.0315}\uparrow$} \\ \cmidrule(r){2-14}
& \multirow{2}{*}{\makecell[c]{New\\Year's Day}} & MSE & 0.0095 & 0.0083 & 0.0044 & 0.0034 & 0.0034 & 0.0036 & 0.0032 & 0.0039 & 0.0033 & \textcolor{darkgreen}{\textbf{0.0030}} & \textcolor{blue}{$\textbf{0.0016}\uparrow$} \\
                       &   & MAE & 0.0841 & 0.0761 & 0.0453 & 0.0388 & 0.0363 & 0.0416 & 0.0372 & 0.0430 & 0.0361 & \textcolor{darkgreen}{\textbf{0.0342}} & \textcolor{blue}{$\textbf{0.0260}\uparrow$} \\
\bottomrule
\end{tabularx}
\footnotesize
\textit{Note:} The best results are marked in blue bold. And the second is marked in green bold.
\end{table*}
\begin{table*}[!t]
\centering
\caption{Performance Comparison with Different Prediction Window Sizes on Traffic}
\label{tab:Comparison Traffic}
\small
\scriptsize 
\belowrulesep=0pt
\aboverulesep=0pt
\setlength{\tabcolsep}{5pt}
\begin{tabularx}{\textwidth}{cc|ccccccccccc}
\toprule
\multirow{3}{*}{\makecell[c]{Predictive\\Size}} & \multirow{3}{*}{Metrics} & \multicolumn{2}{c|}{RNN-based} & \multicolumn{3}{c|}{Transformer-based} & \multicolumn{3}{c|}{Graph-based} & \multicolumn{1}{c|}{CNN-based} & \multicolumn{1}{c|}{Multi-scale} & \multirow{3}{*}{\textbf{Ours}}\\
\cmidrule(r){3-12}  \cmidrule(lr{0pt}){11-12} 
 &  & \multicolumn{1}{c}{LSTM} & \multicolumn{1}{c|}{GRU} & \makecell[c]{Transformer\\NeurIPS'2017} & \makecell[c]{Pyraformer\\ICLR'2022} &  \multicolumn{1}{c|}{\makecell[c]{iTransformer\\ICLR'2024}}  & \makecell[c]{ASTGCN\\AAAI'2019} & \makecell[c]{ST-GDN\\AAAI'2021} & \multicolumn{1}{c|}{\makecell[c]{Dstagnn\\ICML'2022}} & \multicolumn{1}{c|}{\makecell[c]{TimesNet\\ICLR'2023}} & \multicolumn{1}{c|}{\makecell[c]{TimeMixer\\ICLR'2024}} &  \\
\midrule
\multirow{2}{*}{1} & MSE & 0.0058 & 0.0048 & 0.0042 & 0.0050 & \textcolor{darkgreen}{\textbf{0.0023}} & \textcolor{darkgreen}{\textbf{0.0023}} & 0.0033 & 0.0033 & 0.0030 & 0.0029 & \textcolor{blue}{$\textbf{0.0015}\uparrow$} \\ 
                  & MAE & 0.0457 & 0.0415 & 0.0322 & 0.0364 & \textcolor{darkgreen}{\textbf{0.0213}} & 0.0230 & 0.0295 & 0.0326 & 0.0275 & 0.0255 & \textcolor{blue}{$\textbf{0.0174}\uparrow$} \\
\midrule
\multirow{2}{*}{4} & MSE & 0.0095 & 0.0079 & 0.0057 & 0.0056 & \textcolor{darkgreen}{\textbf{0.0035}} & 0.0040 & 0.0040 & 0.0045 & 0.0036 & 0.0047 & \textcolor{blue}{$\textbf{0.0031}\uparrow$} \\
                   & MAE & 0.0885 & 0.0681 & 0.0411 & 0.0393 & \textcolor{darkgreen}{\textbf{0.0263}} & 0.0331 & 0.0033 & 0.0392 & 0.0298 & 0.0326 & \textcolor{blue}{$\textbf{0.0249}\uparrow$} \\
\midrule
\multirow{2}{*}{8} & MSE & 0.0100 & 0.0091 & 0.0053 & 0.0054 & 0.0041 & 0.0047 & \textcolor{darkgreen}{\textbf{0.0040}} & 0.0047 &  \textcolor{blue}{\textbf{0.0039}} & 0.0051 & \textcolor{blue}{\textbf{0.0039}} \\
                   & MAE & 0.1049 & 0.0874 & 0.0381 & 0.0385 & \textcolor{blue}{\textbf{0.0297}} & 0.0365 & 0.0315 & 0.0396 & \textcolor{darkgreen}{\textbf{0.0310}} & 0.0362 & 0.0315 \\
\bottomrule
\end{tabularx}
\footnotesize
\textit{Note:} The best result is marked in blue bold. And the second is marked in green bold.
\end{table*}

\subsection{Evaluation Metrics}
This paper adopts two evaluation metrics, MAE and MSE. All errors in the test set are averaged and compared firstly. Furthermore, for GZN from the railway station, the prediction error during specific time periods (i.e., evening rush hours, weekends, and holidays) is further analyzed.

\subsection{Baseline Methods}
For this study, there are ten baseline methods used:
\begin{itemize}
\item{\textbf{LSTM}}\cite{hochreiter1997long}: Long short-term memory, a variant of RNN.
\item{\textbf{GRU}}\cite{chung2014empirical}: Gated recurrent unit, a variant of RNN.
\item{\textbf{Transformer}}\cite{vaswani2017attention}: The self-attention mechanism is applied for long-sequence modeling.
\item{\textbf{Pyraformer}}\cite{liu2021pyraformer}: A transformer-based method that applies a novel pyramid attention mechanism to bridge the gap between capturing long-distance dependencies and achieving low temporal and spatial complexity.
\item{\textbf{iTransformer}}\cite{liu2023itransformer}: A transformer-based method that using self-attention to capture correlations between variables and the feed forward network to form the global representation within the sequence.
\item{\textbf{ASTGCN}}\cite{guo2019attention}: A graph-based framework that combines the attention mechanism.
\item{\textbf{ST-GDN}}\cite{zhang2021traffic}: A graph-based framework that learns both the local and global dependencies and captures multi-level temporal dynamics. 
\item{\textbf{Dstagnn}}\cite{lan2022dstagnn}: A graph-based framework that excavates spatial associations directly from the historical traffic flow data.
\item{\textbf{TimesNet}}\cite{wu2022timesnet}: A convolution-based architecture that converts the 1D time series into 2D structure in terms of intraperiod-variation and interperiod-variation.
\item{\textbf{TimeMixer}}\cite{wang2024timemixer}: An architecture that analyzes temporal variations in a novel view of multi-scale mixing.
\end{itemize}

\subsection{Implementation Details}
This paper employs 12 2-layer TCN blocks and 6 1-layer MaxPooling blocks. The filter size increases linearly with the number of layers in both blocks, expressed as $2i+1$, where $i$ is the index of the layer. The number of components decomposed by CEEMDAN is determined by the complexity of the data itself. For GZN, the number of decomposition channels is 12. Each STDGCRN has one layer and encodes data into spatial-temporal embeddings of size 128. Finally, 2-layer self-attention blocks are used to enhance temporal features. As shown in Fig. \ref{fig:Framework}, except for the second step, which encodes the data for the three temporal enhancement channels, the first and third steps are both learned and processed separately for each node.

During the training, this paper uses an AdamW optimizer with a batch size of 128 and the weight decay is set to 0.0001. The duration of the early stop is 6. The initial learning rate is set to $1\times 10^{-4}$. The parameters in the EPEL function are set to $p=2$ and $q=1$. All experiments are performed on the PyTorch framework using Python language with 1 NVIDIA GeForce RTX 3060. 

\subsection{Comparative experiments}
To better demonstrate the spatial-temporal modeling capabilities of different methods, this section compares the errors on the entire test set from both GZN and Traffic for prediction window sizes of 1, 4, and 8, respectively, setting a historical window size of 24. The quantitative results are are presented in Tables \ref{tab:Comparison GZN} and \ref{tab:Comparison Traffic}. In addition, the daily evening peak period (i.e., 17:00 - 20:00), weekends (i.e., Saturday and Sunday), and around New Year's Day (i.e., from December 30, 2023, to January 3, 2024 ) on GZN are individually selected for error analysis to highlight the performance of different methods. The accuracy of passenger flow forecasting under these abnormal conditions, especially during peak periods, is the main concern of the railway station management and is also the key to improving the service level.

Comparative results over different time periods performed on GZN can be seen in Table \ref{tab:Comparison GZN}. Comparing a variety of advanced methods based on different architectures, except that the MAE error in the evening is slightly lower than that of ST-GDN, MM-STFlowNet obtains the smallest prediction error in experiments during different time periods with different prediction window sizes. The prediction performance of MM-STFlowNet is not only superior to other methods in the whole test set, but also maintains good prediction accuracy in abnormal cases. The comparison results on Traffic are shown in Table \ref{tab:Comparison Traffic} to further prove the effectiveness. MM-STFlowNet with a prediction window size of 8 has a slightly higher MAE compared to iTransformer on Traffic. Among the other predictions, it has the smallest error. Therefore, the better predictive performance is further validated on an urban road dataset, which contains more nodes and samples.
\begin{figure*}[bp]
\centering
\includegraphics[width=6.4in]{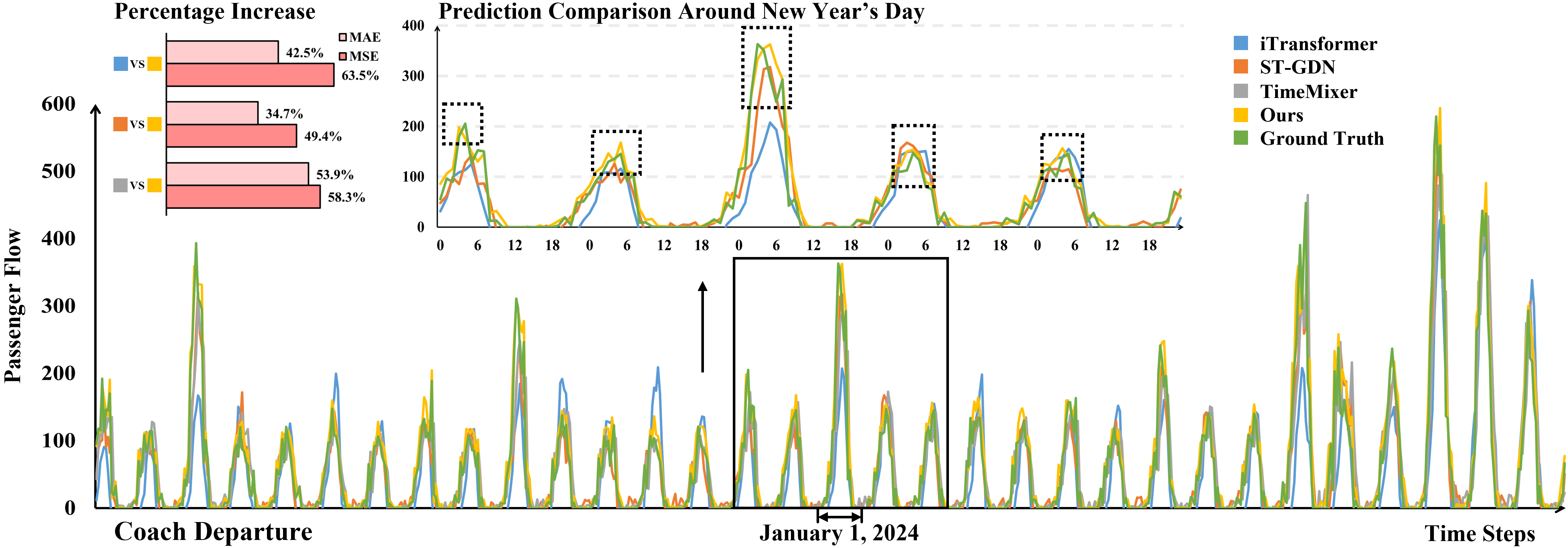}
\caption{Visualization of error comparison in coach departure passenger flow when the prediction window size is 1.}
\label{fig:Coach Comparision}
\end{figure*}
\begin{figure*}[!t]   
\centering
\includegraphics[width=6.4in]{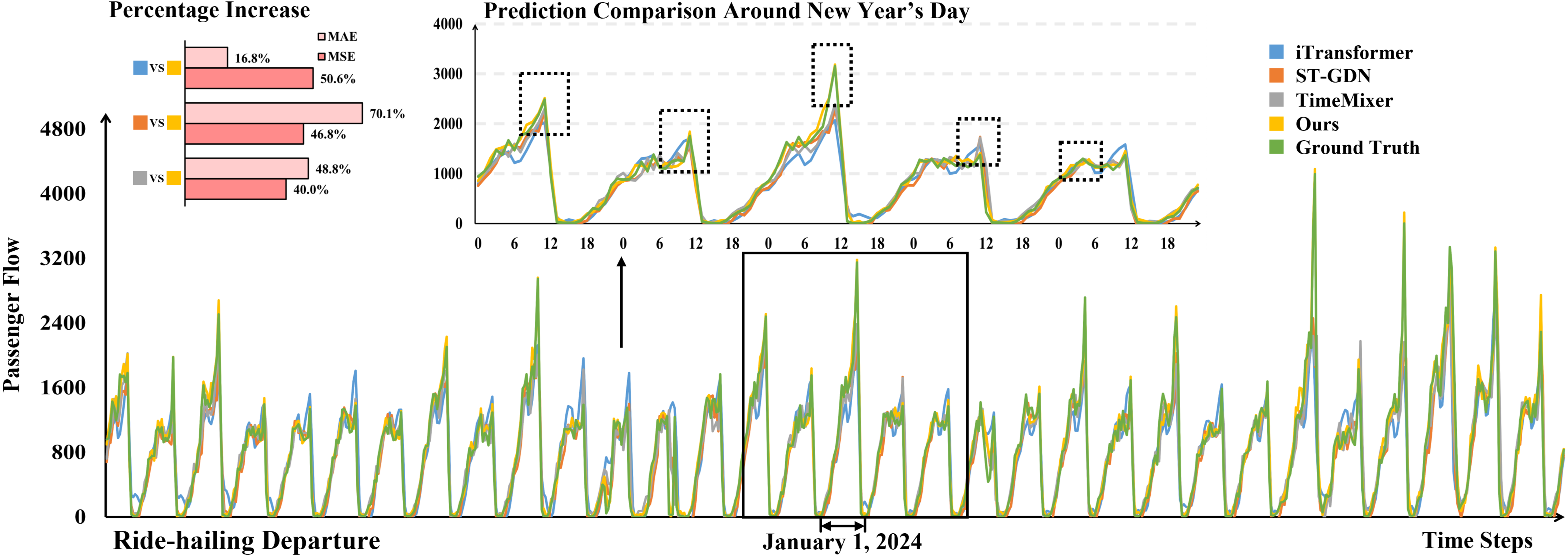}
\caption{Visualization of error comparison for ride-hailing departure passenger flow when the prediction window size is 1.}
\label{fig:Ride-hailing Comparision}
\end{figure*}
\begin{figure*}[!t]
\centering
\includegraphics[width=6.4in]{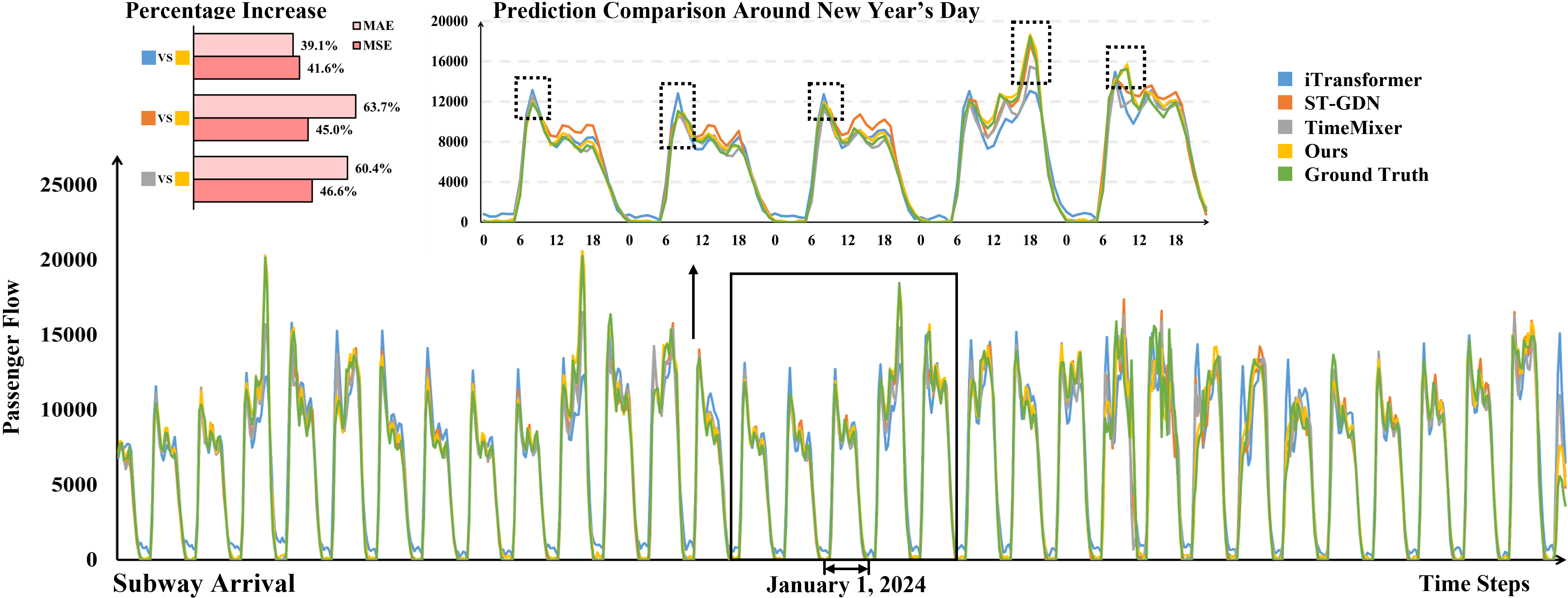}
\caption{Visualization of error comparison for subway arrival passenger flow when the prediction window size is 1.}
\label{fig:Subway Comparision}
\end{figure*}

In addition, three methods (i.e., iTransformer \cite{liu2023itransformer}, ST-GDN \cite{zhang2021traffic} and TimeMixer \cite{wang2024timemixer}) are selected for visual analysis according to performance. Similar to Fig. \ref{fig:Data Display}, three passenger flows (i.e., coach departure, ride-hailing departure and subway arrival) are used to demonstrate the different temporal patterns. The prediction errors for the peak during the New Year's Day period are further visualized, which can be shown from Fig. \ref{fig:Coach Comparision} to Fig. \ref{fig:Subway Comparision}. The performance comparison around New Year's Day provides a more intuitive display, and the statistics on the percentage increase compared to the other three methods is shown at the top. In terms of the overall trend, MM-STFlowNet is closest to the true value. Around New Year's Day, the timing and magnitude of the peak vary significantly for these three passenger flows, which is a forecasting challenge under abnormal conditions. Through the contrast of the spikes highlighted by the dotted black frame in Figs. \ref{fig:Coach Comparision}, \ref{fig:Ride-hailing Comparision} and \ref{fig:Subway Comparision}, MM-STFlowNet approximates the real peak with the minimum error, which can be used as the best basis for taking emergency measures at train stations. For example, on New Year's Eve, the peak passenger flow of ride-hailing departure is predicted to be around 3,200 in Fig. \ref{fig:Ride-hailing Comparision}, which means that more ride-hailing services need to be provided and traffic pressure must be relieved on the main roads leading from the train station to the city center. Then, the number of people returning home to the railway stations after the holiday has increased rapidly. The peak passenger flow arriving by subway is predicted around 18,000 in Fig. \ref{fig:Subway Comparision}. Therefore, the orderly departure of people from the exit of the subway station should be reinforced to prevent serious stampede accidents. Furthermore, around New Year's Day, the percentage improvement compared with other three methods in predicted performance is demonstrated using a histogram in Figs \ref{fig:Coach Comparision}, \ref{fig:Ride-hailing Comparision} and \ref{fig:Subway Comparision}. The maximum increase is 70.1$\%$, and the minimum is 16.8$\%$. The prediction robustness of MM-STFlowNet in abnormal cases is more intuitively demonstrated. 

\begin{figure}[!t]
\centering
\subfloat[Multi-mode arrival flow.]{\includegraphics[width=1.5in]{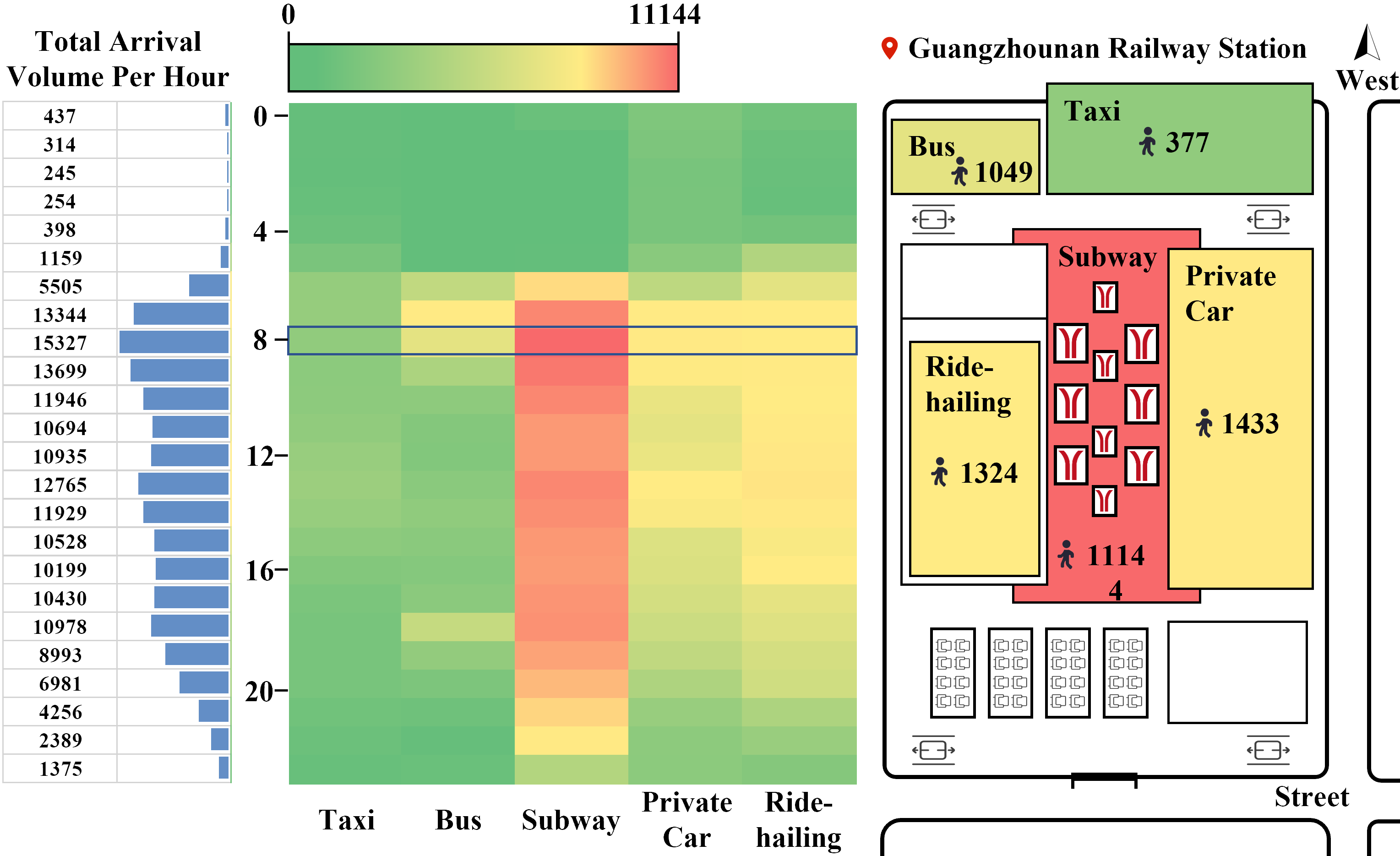}\label{fig:Arrival NYD}}
\hfil
\subfloat[Multi-mode departure flow.]{\includegraphics[width=1.6in]{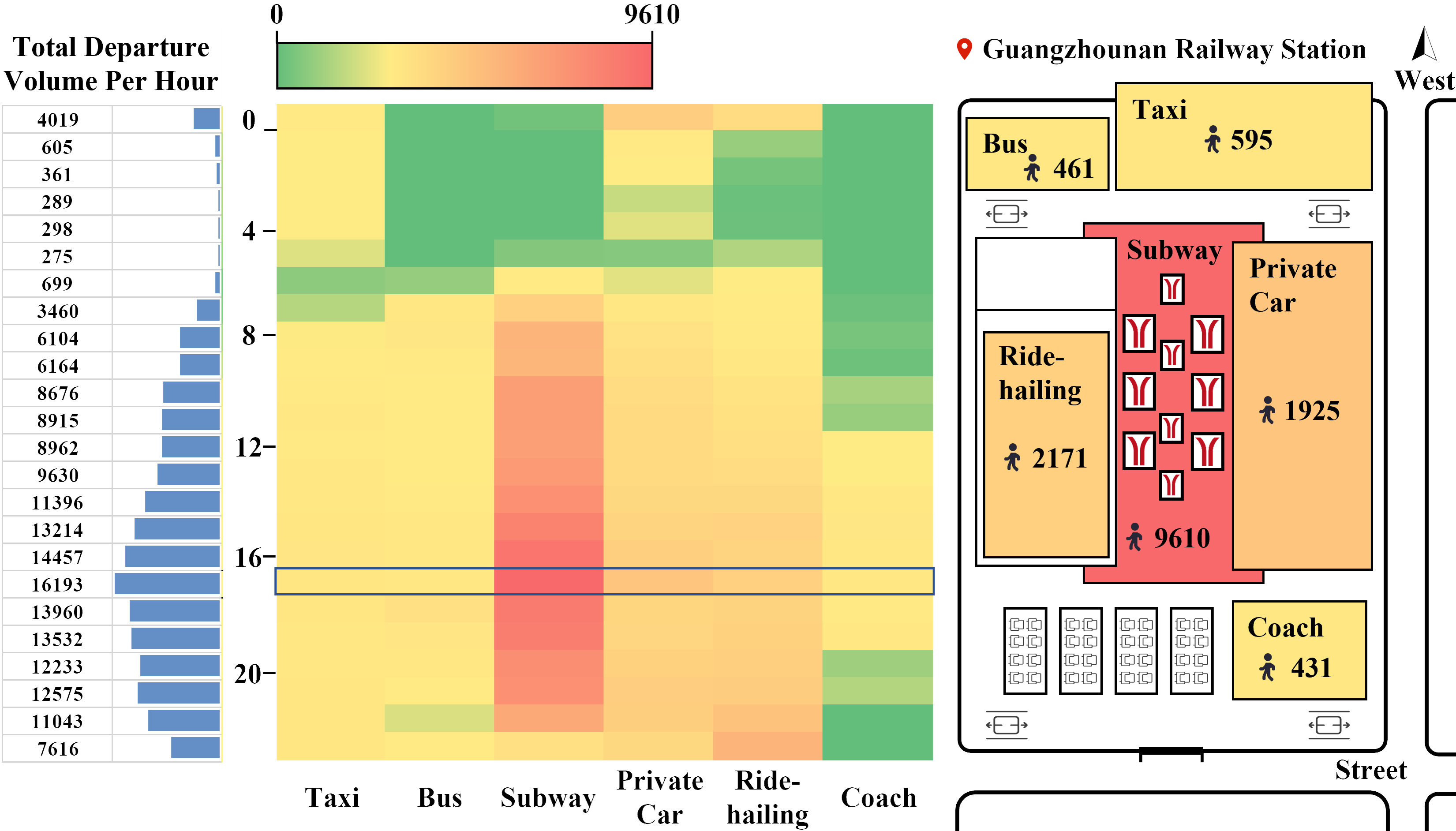}\label{fig:Evacuation NYD}}
\caption{Heat maps and distribution of multi-mode passenger flows on New Year's Day. This paper briefly represents the distribution of each mode of transportation, referring to Guangzhounan Railway Station.}
\label{fig:Heat1}
\end{figure}

Moreover, to demonstrate the superiority of multi-mode refined prediction in detailed analysis, this paper presents the prediction results for New Year's Day using heat maps and exhibits the distribution of passengers at a moment on the train station in Fig. \ref{fig:Heat1}. The total volume of arrivals and departures per hour is also clearly obtained, which is the same as the previous research. Among the predictive changes in both directions (i.e., arrival and departure), passengers prefer to take the subway owing to its better convenience and performance price ratio. And the peak for arriving at the train station by subway is predicted at 8 a.m. in Fig. \ref{fig:Heat1}\subref{fig:Arrival NYD} and the peak for leaving is predicted at 5 p.m. in Fig. \ref{fig:Heat1}\subref{fig:Evacuation NYD}. Based on the above information, on the one hand, the frequency of subway services should be increased during peak hours. At other times, the frequency can be reduced to avoid resource waste. On the other hand, the management should ensure timely dispersal and guidance for arriving passengers and organize alternative transportation for departing passengers to avoid congestion. Unlike predicting the total flow within the station, through multi-mode refined forecasting, staff can be assigned to the entrances and exits of the traffic mode that is about to peak. In this way, emergency measures can be prepared in a timely and adequate manner. In addition, real-time sharing of passenger flow information can be realized, so that each traffic mode can adjust the operation strategy with each other. For example, 1,049 people are predicted to take the bus at 8 a.m. arriving at the train station in Fig. \ref{fig:Heat1}\subref{fig:Arrival NYD}. Due to the early hours, there are often long queues of passengers waiting for the bus. With access to the predictive data, taxi drivers can head to the corresponding bus stops to pick up passengers. At the same time, the predicted number of passengers by bus in the morning is higher than in the afternoon. Some bus routes may need to be dynamically adjusted according to different periods to provide a better experience. Similarly, 2,171 people are predicted to take the ride-hailing service at 5 p.m. in Fig. \ref{fig:Heat1}\subref{fig:Evacuation NYD}. Many passengers cancel their ride-hailing orders because of a long time to wait, which may require more taxis.
\begin{figure}[!t]
\vspace{-0.5cm}
\centering
\subfloat[Multi-mode arrival flow on the weekday.]{\includegraphics[width=1.2in]{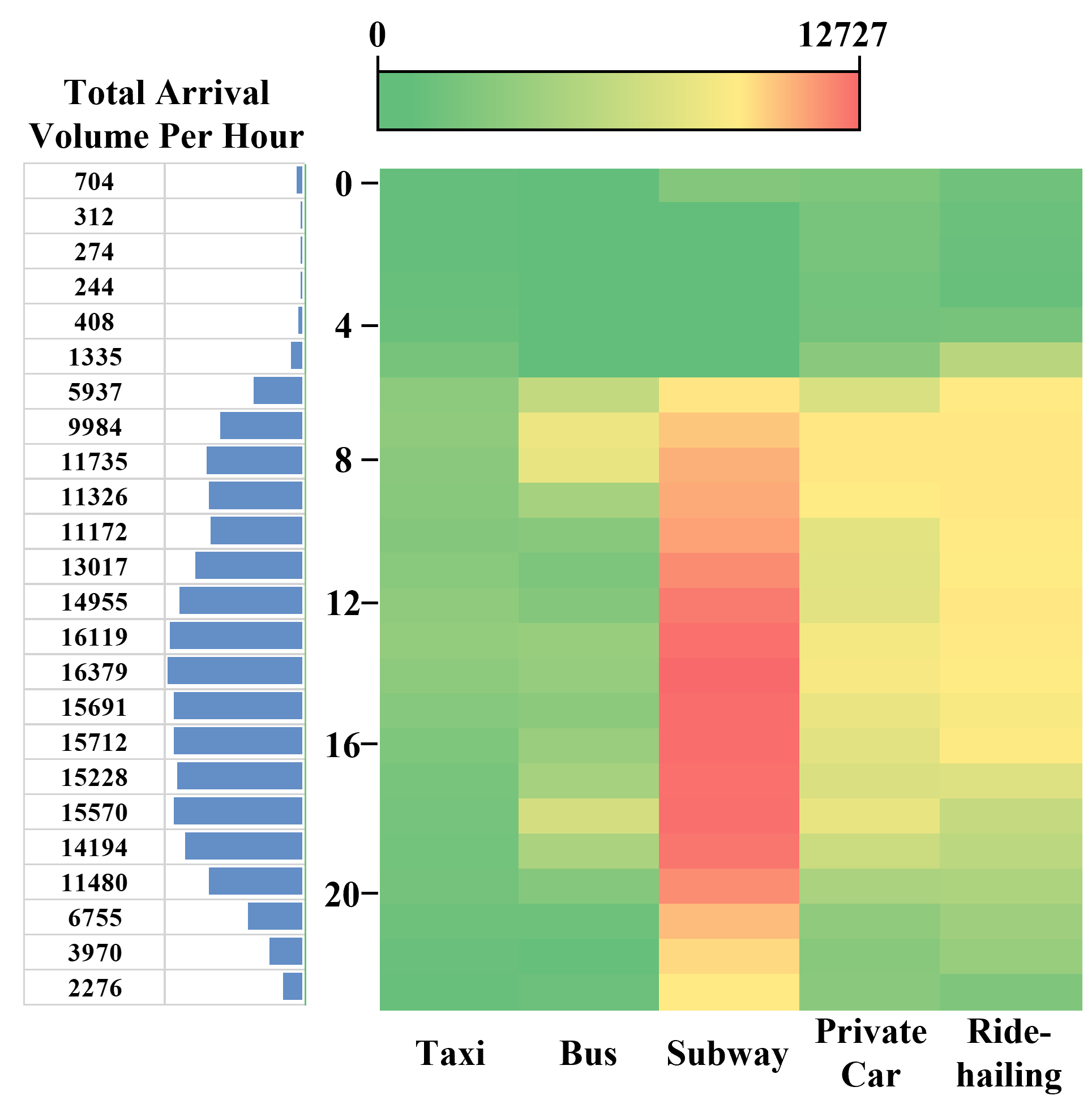}\label{fig:Arrival Workday}}
\hfil
\subfloat[Multi-mode departure flow on the weekday.]{\includegraphics[width=1.38in]{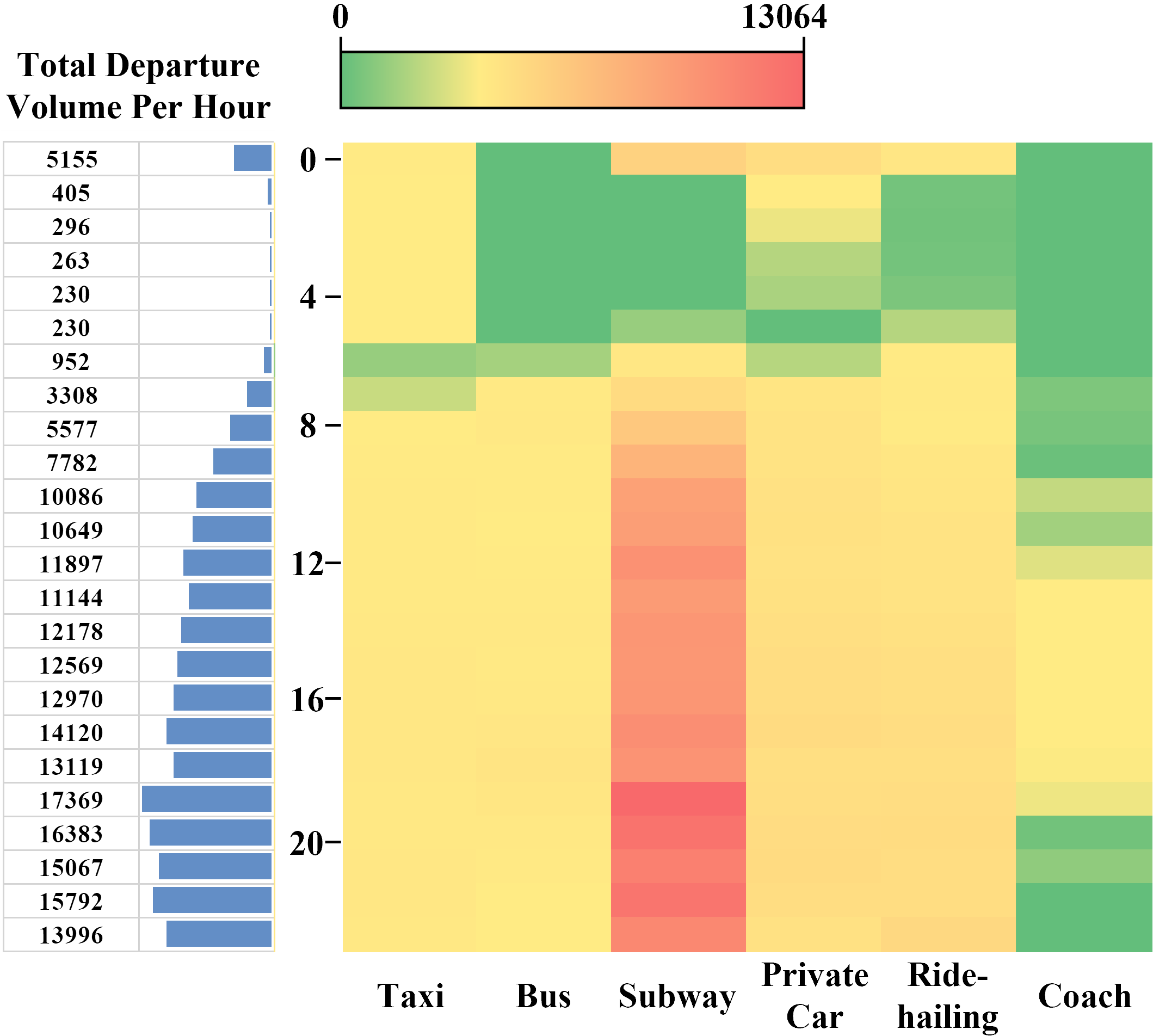}\label{fig:Evacuation Workday}}
\hfil
\subfloat[Multi-mode arrival flow on the weekend.]{\includegraphics[width=1.2in]{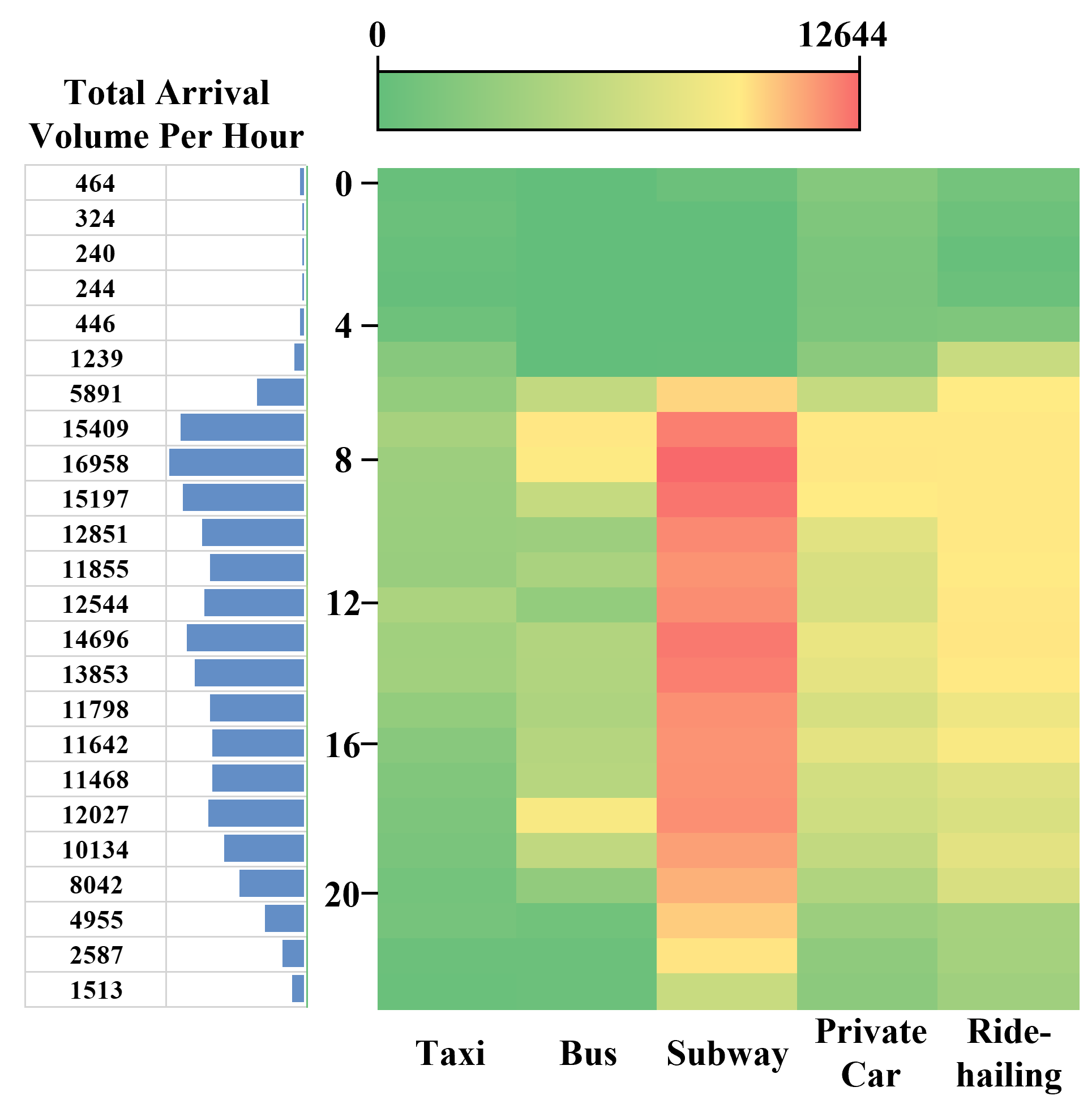}\label{fig:Arrival Weekend}}
\hfil
\subfloat[Multi-mode departure flow on the weekend.]{\includegraphics[width=1.38in]{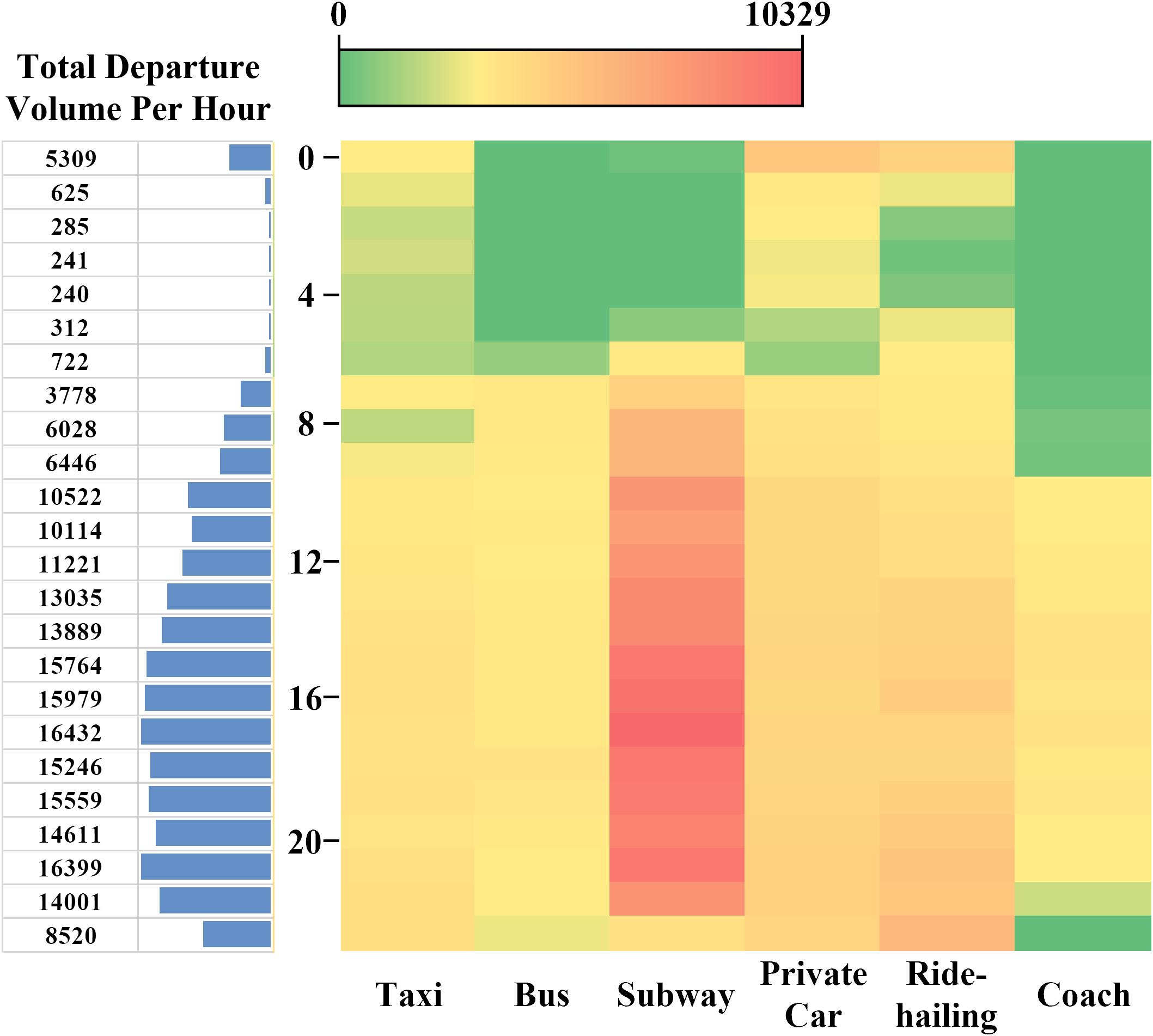}\label{fig:Evacuation Weekend}}
\caption{Heat maps of multi-mode passenger flows on the weekday and weekend. The date of the weekend is January 14, 2024. And the date of the weekday is January 10, 2024.}
\label{fig:Heat2}
\vspace{-0.5cm}
\end{figure}

The daily difference in traffic patterns can be shown in Fig. \ref{fig:Heat2}. Firstly, the peak of passenger flow arriving or leaving on weekends always appears in advance. Secondly, the number of people traveling by individual motor vehicles on weekends increased significantly in the afternoons and evenings. Thirdly, whether on weekdays or weekends, the subway is almost the first choice for people to arrive at the train station, while the modes to leave are more diverse. Based on the above findings, operators can gain insight into traffic differences in advance and take countermeasures in advance.

In summary, refined forecasting provides insight into not only the future distribution of passenger flows, but also the variability of traffic trends in each mode, which is the practical value that traditional forecasting methods cannot reach.
\begin{table*}[bp]
\centering
\caption{Predictive Performance of the Proposed method with Different Historical Window Sizes on GZN}
\label{tab:Window Sizes}
\small
\scriptsize 
\belowrulesep=0pt
\aboverulesep=0pt
\setlength{\tabcolsep}{6.3pt}
\begin{tabularx}{\textwidth}{cc|cccccccccccccc}
\toprule
\multirow{2}{*}{\makecell[c]{Predictive\\Size}} & \multirow{2}{*}{Metrics} & \multicolumn{14}{c}{Historical Window Size} \\
\cmidrule(r){3-16}
 &  & 8 & 12 & 16 & 20 & \textbf{24} & 28 & \textbf{32} & 36 & 48 & 64 & 80 & 96 & \textbf{128} & 256 \\
\midrule
\multirow{2}{*}{1} & MSE & 0.0039 & 0.0022 & 0.0019  & 0.0015 & \textcolor{blue}{\textbf{0.0012}} & 0.0021 & 0.0018 & 0.0023 & 0.0018 & 0.0026 & 0.0022 & 0.0025 & 0.0034 & 0.0035 \\ 
                  & MAE & 0.0445 & 0.0300 & 0.0284  & 0.0247 & \textcolor{blue}{\textbf{0.0222}} & 0.0292 & 0.0277 & 0.0315 & 0.0270 & 0.0334 & 0.0307 & 0.0325 & 0.0369 & 0.0378 \\
\midrule
\multirow{2}{*}{4} & MSE & 0.0053 & 0.0044 & 0.0041  & 0.0039 & 0.0022 & 0.0022 & \textcolor{blue}{\textbf{0.0021}} & 0.0024 & 0.0024 & 0.0023 & 0.0024 & 0.0024 & 0.0028 & 0.0027 \\
                   & MAE & 0.0444 & 0.0417 & 0.0405  & 0.0397 & 0.0297 & 0.0311 & \textcolor{blue}{\textbf{0.0293}} & 0.0300 & 0.0307 & 0.0294 & 0.0304 & 0.0299 & 0.0333 & 0.0331 \\
\midrule
\multirow{2}{*}{8} & MSE & 0.0072 & 0.0058 & 0.0051  & 0.0032 & 0.0025 & 0.0025 & 0.0026 & 0.0026 & 0.0024 & 0.0024 & 0.0023 & 0.0023 & \textcolor{blue}{\textbf{0.0021}} & 0.0022 \\
                   & MAE & 0.0510 & 0.0467 & 0.0435  & 0.0318 & 0.0306 & 0.0293 & 0.0309 & 0.0306 & 0.0287 & 0.0285 & 0.0282 & 0.0283 & \textcolor{blue}{\textbf{0.0266}} & 0.0286 \\
\bottomrule
\end{tabularx}
\footnotesize
\textit{Note:} The best result is marked in blue bold.
\end{table*}

\subsection{Optimal History Window Size}
At train stations, the most urgent need of the authorities is to accurately obtain the future flow pressure of passengers. So relevant emergency measures can be made in time. The predictive ability is affected differently depending on the size of the historical window. This paper also explores the optimal history window sizes when the prediction window size is 1, 4 and 8 using MM-STFlowNet on GZN. The results are presented in Table \ref{tab:Window Sizes}.

In Table \ref{tab:Window Sizes}, as the difficulty of forecasting continues to increase, so does the need for historical data. The best historical window sizes for predicting the next 1h, 4h, and 8h passenger traffic data are 24, 32, and 128, respectively. When forecasting passenger volume for the next 1h, it is best to provide a complete period (i.e., 24h), which maybe includes the most appropriate spatial-temporal features required. This is also the most accurate way to predict reference data in terms of error. But it's also the least difficult task compared to other long-sequence predictions. For larger prediction sizes, a short history may result in insufficient learning of temporal patterns, while a longer one may cause redundancy and confusion in information. In addition to the modeling method, it is also critical to select the appropriate historical length to capture the adequate spatial-temporal features for long-sequence prediction tasks\cite{shi2024scaling,edwards2024scaling}. These results in Table \ref{tab:Window Sizes} demonstrate the importance of selecting the appropriate historical window sizes tailored to the specific prediction.

\subsection{Ablation Studies}
To further evaluate the independent contribution of each module, different models in ablation studies preformed on GZN are constructed as follows: 
\begin{itemize}
\item{\textbf{ND}}: Series decomposition based on CEEMDAN is removed.
\item{\textbf{NH}}: Historical enhancement based on TCN blocks is removed.
\item{\textbf{NP}}: Peak amplification based on MaxPooling blocks is removed.
\item{\textbf{NDH}}: Both series decomposition and historical enhancement are removed.
\item{\textbf{NDP}}: Both series decomposition and peak amplification are removed.
\item{\textbf{NHP}}: Both historical enhancement and peak amplification are removed.
\item{\textbf{NDHP}}: Series decomposition, historical enhancement and peak amplification are removed.
\item{\textbf{NDG}}: Dynamic graph convolution is removed. Embeddings are encoded by GRU.
\item{\textbf{NCA}}: The channel attention fusion is removed. Different embeddings are added directly.
\item{\textbf{NSA}}: The self-attention enhancement is removed. The embeddings fused by the channel attention mechanism are projected directly to the results.
\end{itemize}

\begin{figure}[H]
\centering
\includegraphics[width=2.7in]{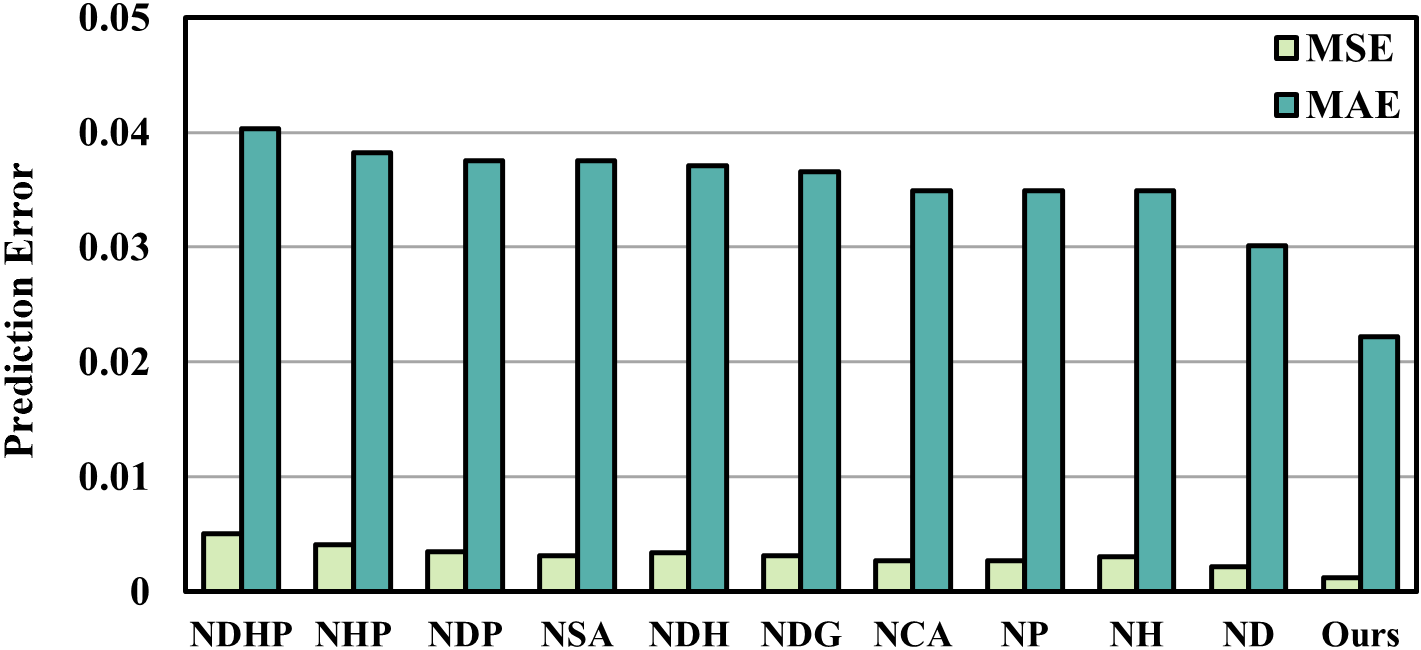}
\caption{Ablation experiments on different variant models.}
\label{fig:Ablation Models}
\end{figure}

All experiments are controlled to use 24h historical data to predict traffic status for the next 1h. The errors of different models are shown in Fig. \ref{fig:Ablation Models}. On GZN, MM-STFlowNet is better than other variant models. Among the three temporal feature processing methods, missing historical enhancement and peak amplification have almost the same significant impact on the final prediction performance. And the dynamic graph convolution that capture spatial dependencies has a greater impact than all temporal operations, demonstrating its effectiveness. In addition, feature fusion and enhancement after spatial-temporal encoding are also very important.

\begin{figure}[H]
\centering
\includegraphics[width=2.7in]{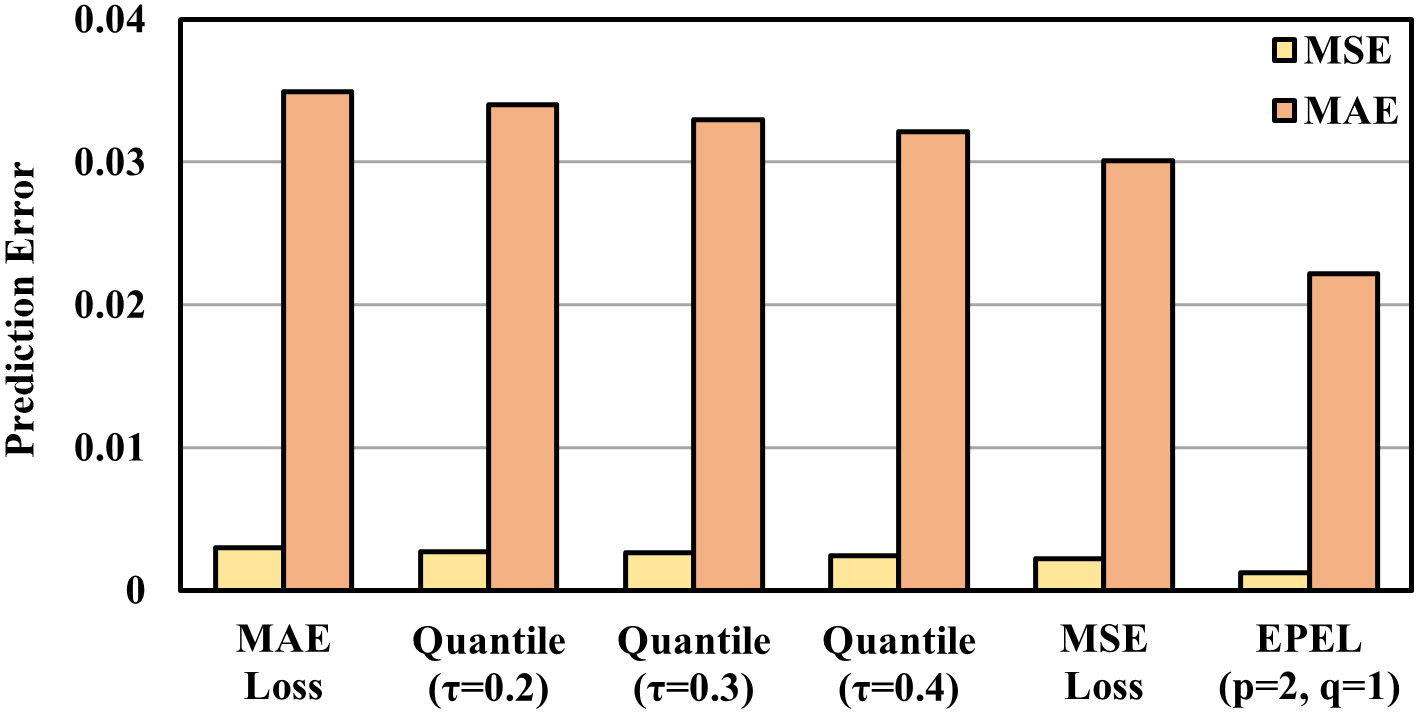}
\caption{Ablation experiments on different loss functions.}
\label{fig:Ablation Loss}
\end{figure}

This paper also verifies the effectiveness of the EPEL function. The results can be seen in Fig. \ref{fig:Ablation Loss}. The prediction error of MM-STFlowNet is the smallest through EPEL. Due to the more penalties for outliers, the MSE and quantile loss functions achieve relatively better results on GZN, which is also the main basis for the proposed EPEL function. Different degrees of punishment can be carried out according to the size of the true value, so EPEL is more targeted and the training effect is better. 
\vspace{-0.5cm}

\section{Conclusion and Future Development}
\label{sec:Conclusion and Future Development}
This paper presents a comprehensive study on passenger flow forecasting in large-scale transportation hubs, identifying a critical gap in existing research: the lack of detailed exploration of interactions between various transportation modes. To address this, we introduce MM-STFlowNet, a sophisticated prediction framework grounded in dynamic spatial-temporal modeling. This approach not only enhances the accuracy of predicting passenger volumes across different collection and distribution modes, but also provides a granular view of passenger flow distribution within transportation hubs. Our methodology includes advanced temporal processing strategies, such as sequence decomposition, historical enhancement, and peak amplification, to effectively manage data peaks and high volatility. Furthermore, we develop the Spatial-Temporal Dynamic Graph Convolutional Recurrent Network (STDGCRN) to capture intricate local and global dependencies in passenger flow dynamics, supplemented by an adaptive channel attention mechanism for extracting critical features across diverse prediction tasks. The integration of a self-attention mechanism further refines our model by incorporating the impact of various external factors on historical and future data. Extensive experiments on real-world datasets from Guangzhounan Railway Station validate that MM-STFlowNet achieves state-of-the-art prediction performance, particularly during peak periods, providing substantial insight for transportation hub management.

Looking ahead, future research should aim to improve the granularity of predictions by incorporating data at the level of individual entrances and exits for each transportation mode. This refinement poses substantial challenges in terms of data acquisition and significantly increases the complexity of prediction models. However, achieving this level of detail promises to substantially improve predictive accuracy and practical applicability. Furthermore, integrating multi-mode refined prediction with multi-perspective tracking tasks holds considerable potential for advancing the monitoring and prediction of passenger flows in large-scale transportation hubs, thereby contributing to more effective management and resource allocation strategies.

\bibliographystyle{IEEEtran}
\bibliography{Paper}
\begin{IEEEbiography}[{\includegraphics[width=1in,height=1.25in,clip,keepaspectratio]{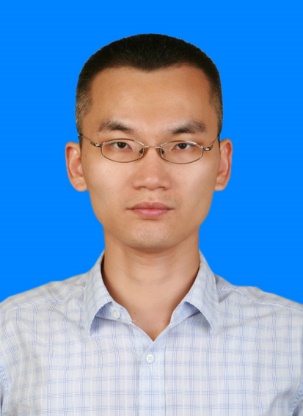}}]{Ronghui Zhang}
received a B.Sc. (Eng.) from the Department of Automation Science and Electrical Engineering, Hebei University, Baoding, China, in 2003, an M.S. degree in Vehicle Application Engineering from Jilin University, Changchun, China, in 2006, and a Ph.D. (Eng.) in Mechanical \& Electrical Engineering from Changchun Institute of Optics, Fine Mechanics and Physics, the Chinese Academy of Sciences, Changchun, China, in 2009. After finishing his post-doctoral research work at INRIA, Paris, France, in February 2011, he is currently an Associate Professor with the Research Center of Intelligent Transportation Systems, School of intelligent systems engineering, Sun Yat-sen University, Guangzhou, Guangdong 510275, P.R.China. His current research interests include computer vision, intelligent control and ITS.
\end{IEEEbiography}
\begin{IEEEbiography}[{\includegraphics[width=1in,height=1.25in,clip,keepaspectratio]{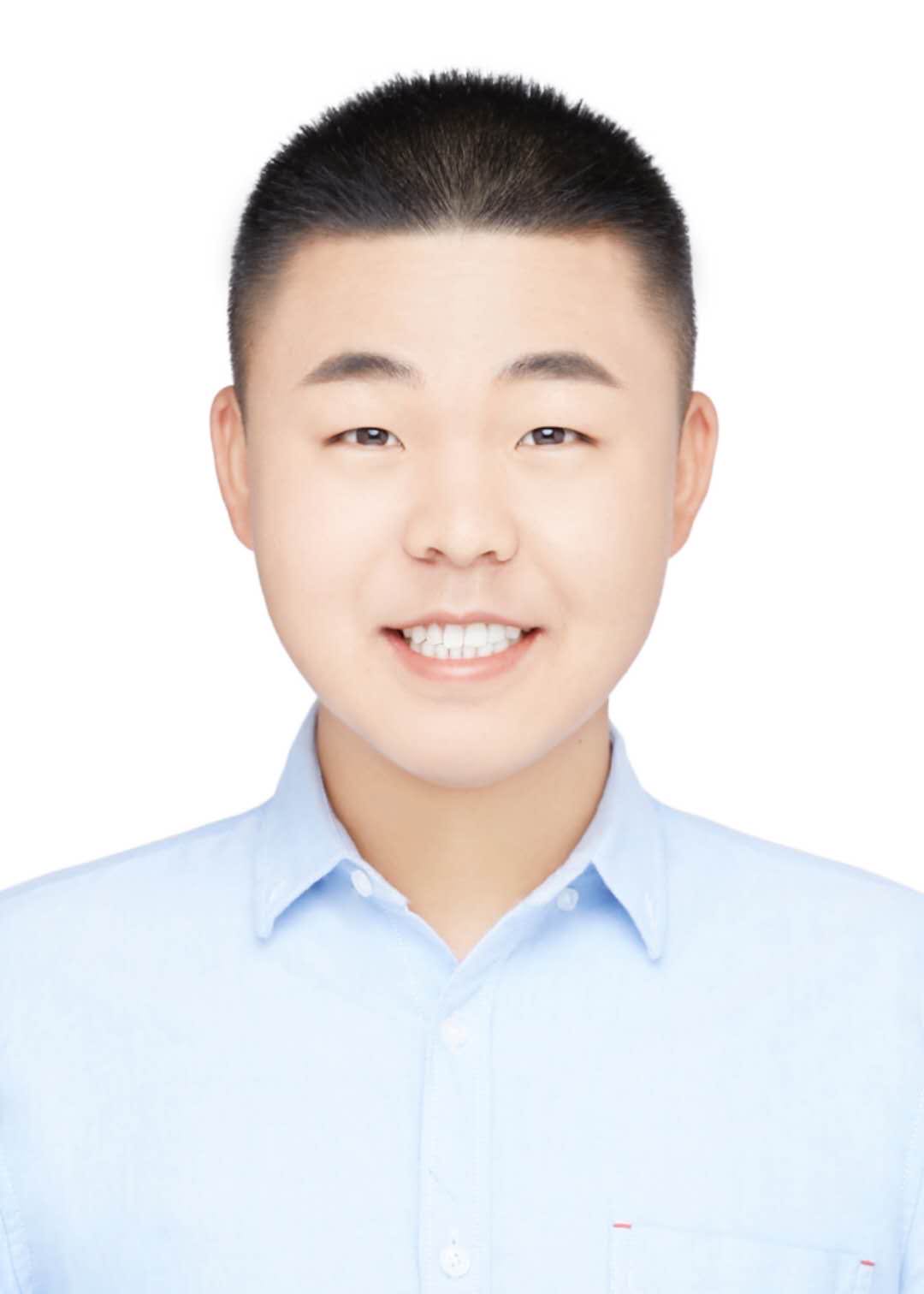}}]{Wenbin Xing}
received his B.S. degree in Computer Science and Technology from Wuhan University of Technology, Wuhan, China, in 2024. He is pursuing his master's degree at Sun Yat-sen University in Shenzhen, China. His current research interests include autonomous driving, traffic big data, deep learning, and spatial-temporal modeling.
\end{IEEEbiography}
\begin{IEEEbiography}[{\includegraphics[width=1in,height=1.25in,clip,keepaspectratio]{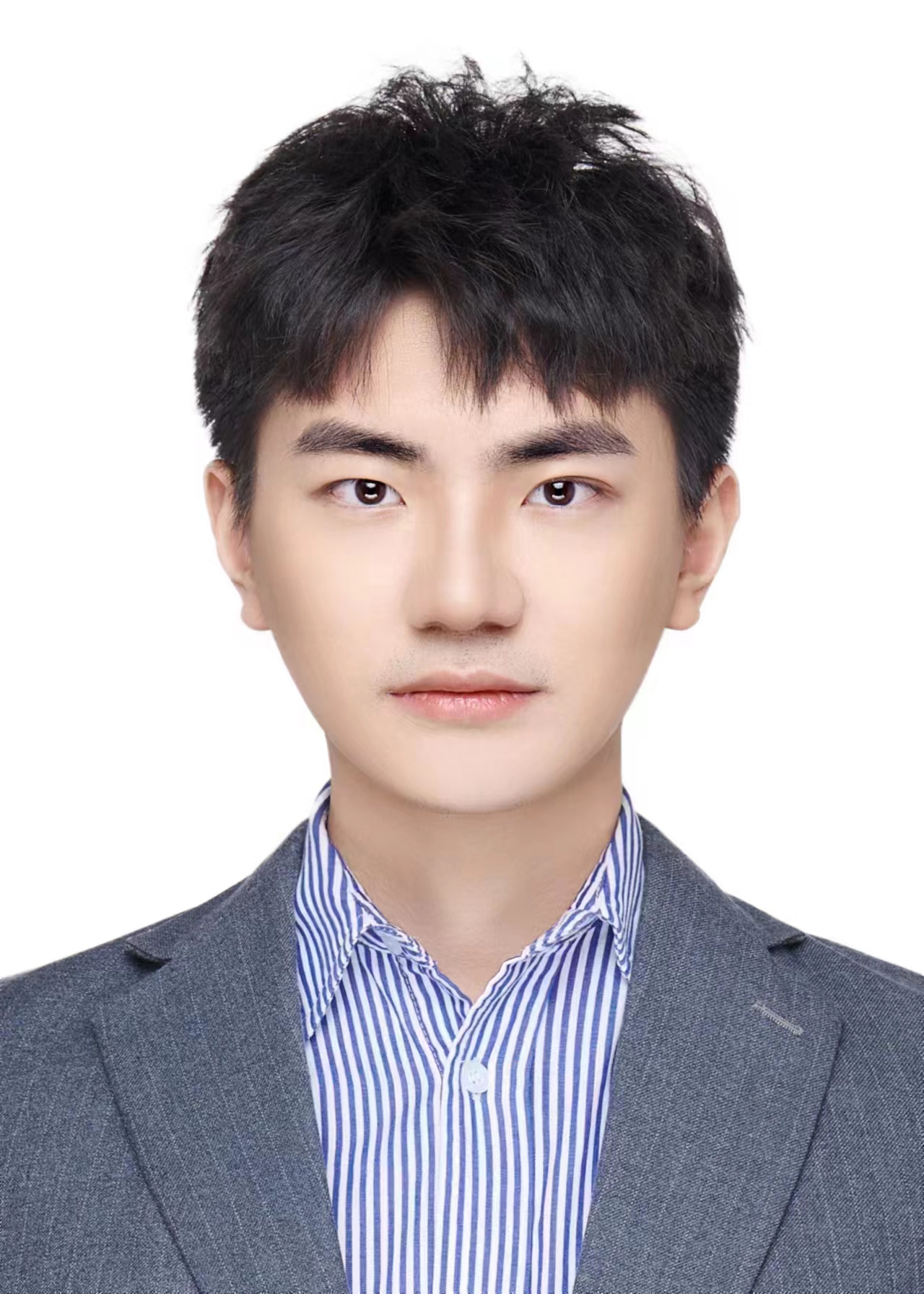}}]{Mengran Li}
received the M.S. degree in Control Science and Engineering from the School of Information Technology at Beijing University of Technology, Beijing, China, in 2023. He is currently pursuing a Ph.D. degree at Guangdong Key Laboratory of Intelligent Transportation System, School of Intelligent Systems Engineering, Sun Yat-sen University, Guangzhou, Shenzhen, China. His research interests include big data and artificial intelligence.
\end{IEEEbiography}
\begin{IEEEbiography}[{\includegraphics[width=1in,height=1.25in,clip,keepaspectratio]{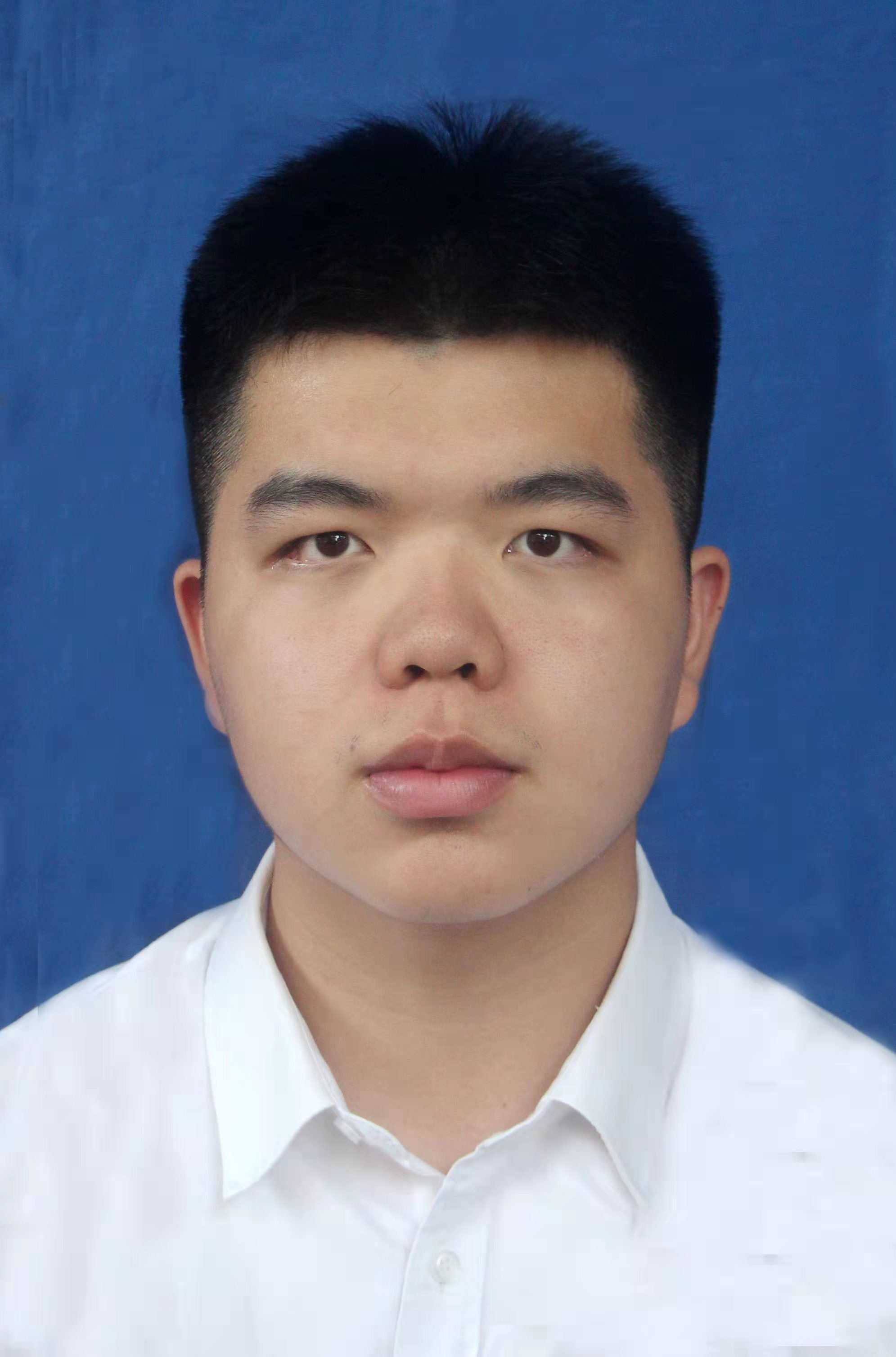}}]{Zihan Wang}
received his B.S. degree in Traffic Engineering from Chang’an University, Xian, China, in 2021. He is currently working towards his master’s degree at Sun Yat-sen University, Guangzhou, China. His current research interests include autonomous driving and vehicle-road collaboration, deep learning and computer vision.
\end{IEEEbiography}
\begin{IEEEbiography}[{\includegraphics[width=1in,height=1.25in,clip,keepaspectratio]{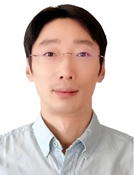}}]{Junzhou Chen}
received the B.S. degree in computer science and applications and the M.Eng. degree in software engineering from Sichuan University, Chengdu, China, in 2005 and 2002, respectively, and the Ph.D. degree in computer science and engineering from The Chinese University of Hong Kong, Hong Kong, in 2008. Between 2009 and 2019, he was a Lecturer and later as an Associate Professor with the School of Information Science and Technology, Southwest Jiaotong University, Chengdu, China. He is currently an Associate Professor with the School of Intelligent Systems Engineering, Sun Yat-sen University, Guangzhou, China. His research interests include computer vision, machine learning, intelligent transportation systems, mobile computing, and medical image processing.
\end{IEEEbiography}
\begin{IEEEbiography}[{\includegraphics[width=1in,height=1.5in,clip,keepaspectratio]{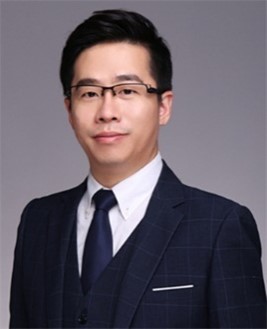}}]{Xiaolei Ma} (Senior Member, IEEE) earned his Ph.D. degree from the University of Washington, Seattle, WA, USA, in 2013. He currently serves as a Professor at the School of Transportation Science and Engineering, Beihang University, China. His research focuses on public transit operations and planning, transportation big data analytics, and the integration of transportation and energy. Dr. Ma is an associate editor for the IEEE Transactions on Intelligent Transportation Systems and IET Intelligent Transport Systems. He also contributes as an editorial member to several journals, including Transportation Research Part C and D, Computers, Environment and Urban Systems, and the Journal of Intelligent Transportation Systems. Dr. Ma is a Fellow of the IET and a Senior Member of the IEEE.
\end{IEEEbiography}
\begin{IEEEbiography}[{\includegraphics[width=1in,height=1.5in,clip,keepaspectratio]{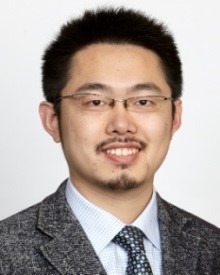}}]{Zhiyuan Liu} (Senior Member, IEEE) received his Ph.D. degree in Transportation Engineering in 2011 from National University of Singapore. He is currently a professor at School of Transportation in Southeast University, and director of the Research Center for Complex Transport Networks. His research interests include transport network modelling, public transport, and intelligent transport system. In these areas, Dr. Liu has published over 100 journal papers.
\end{IEEEbiography}
\begin{IEEEbiography}[{\includegraphics[width=1in,height=1.5in,clip,keepaspectratio]{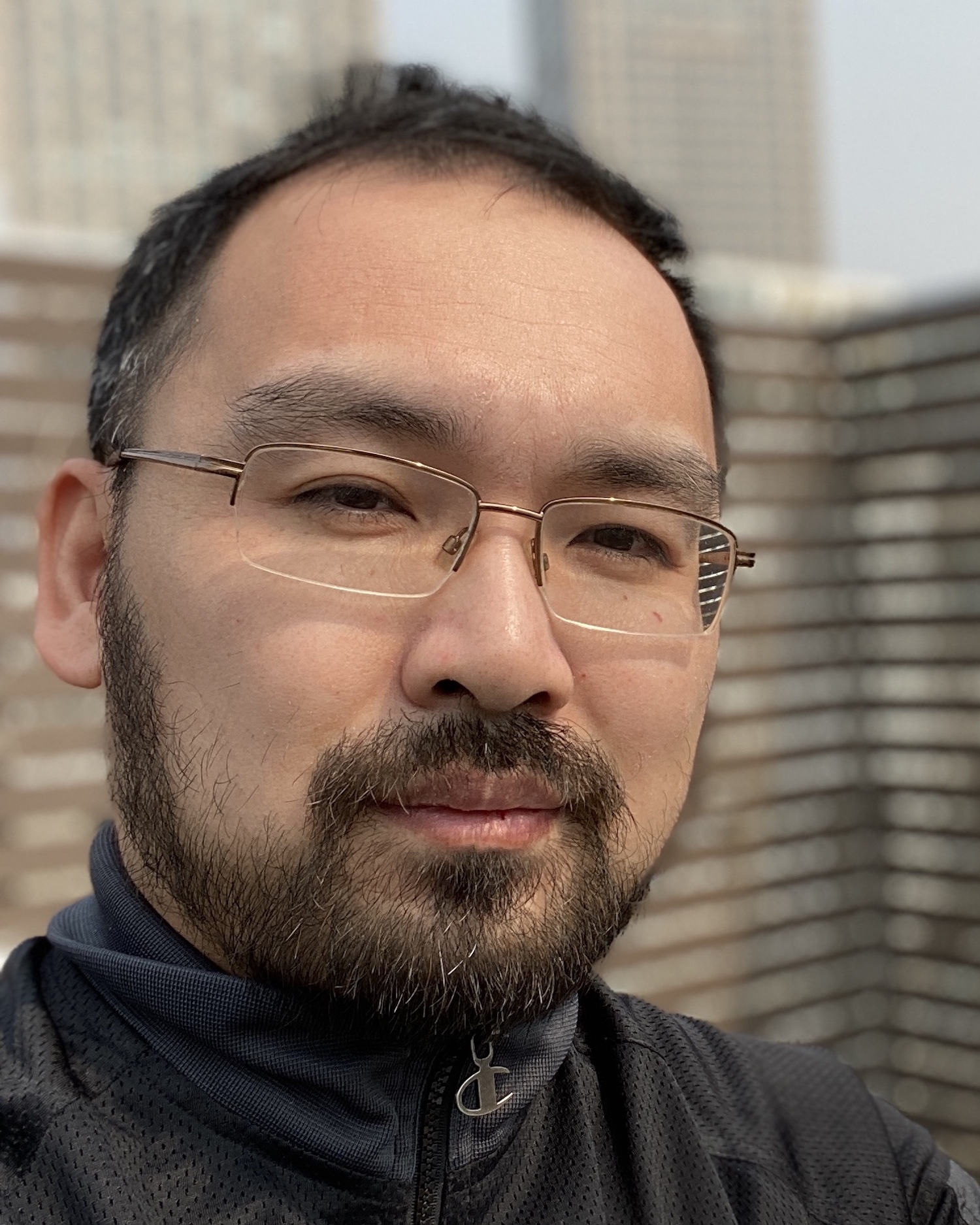}}]{Zhengbin He} (Senior Member, IEEE) is currently a research scientist with Laboratory for Information \& Decision Systems, Massachusetts Institute of Technology, Cambridge, MA, USA. His research interests include traffic flow theory, urban mobility, and sustainable transportation. He is an Associate Editor for IEEE TRANSACTIONS ON INTELLIGENT TRANSPORTATION SYSTEMS, a Handling Editor of Transportation Research Record, and an Editorial Advisory Board Member for Transportation Research Part C. For more information visit the link (https://www.GoTrafhicGo.com).
\end{IEEEbiography}
\end{document}